\documentclass{article}

\usepackage{PRIMEarxiv}
\usepackage[utf8]{inputenc} 
\usepackage[T1]{fontenc}    
\usepackage{hyperref}       
\usepackage{url}            
\usepackage{booktabs}       
\usepackage{amsfonts}       
\usepackage{amsmath}
\usepackage{nicefrac}       
\usepackage{microtype}      
\usepackage{lipsum}
\usepackage{fancyhdr}       
\usepackage{graphicx,subfig}       
\graphicspath{{media/}}     
\usepackage{adjustbox}
\usepackage{cleveref}

\title{Leveraging Representations from Intermediate Encoder-Blocks for Synthetic Image Detection
\thanks{\textit{\underline{Citation}}: 
\textbf{Koutlis, C., \& Papadopoulos, S. (2024). Leveraging Representations from Intermediate Encoder-Blocks for Synthetic Image Detection. In European Conference on Computer Vision. Cham: Springer Nature Switzerland.}} 
}

\author{
  Christos Koutlis, Symeon Papadopoulos \\
  CERTH-ITI \\
  Thessaloniki, Greece\\
  \texttt{\{ckoutlis, papadop\}@iti.gr} \\
}

\begin{document}
\maketitle

\begin{abstract}
The recently developed and publicly available synthetic image generation methods and services make it possible to create extremely realistic imagery on demand, raising great risks for the integrity and safety of online information.
  State-of-the-art Synthetic Image Detection (SID) research has led to strong evidence on the advantages of feature extraction from foundation models. However, such extracted features mostly encapsulate high-level visual semantics instead of fine-grained details, which are more important for the SID task. On the contrary, shallow layers encode low-level visual information. 
  In this work, we leverage the image representations extracted by intermediate Transformer blocks of CLIP's image-encoder via a lightweight network that maps them to a learnable forgery-aware vector space capable of generalizing exceptionally well.
  We also employ a trainable module to incorporate the importance of each Transformer block to the final prediction. 
  Our method is compared against the state-of-the-art by evaluating it on 20 test datasets and exhibits an average +10.6\% absolute performance improvement. Notably, the best performing models require just a single epoch for training ($\sim$8 minutes). Code available at \href{https://github.com/mever-team/rine}{https://github.com/mever-team/rine}.
\end{abstract}


\section{Introduction}
The advancements in the field of synthetic content generation disrupt digital media communications, and pose new societal and economic risks \cite{karnouskos2020artificial}. Generative Adversarial Networks (GAN) \cite{goodfellow2014generative}, the variety of successor generative models \cite{iglesias2023survey} and the latest breed of Diffusion models \cite{croitoru2023diffusion} are capable of producing highly realistic images that often deceive humans \cite{lu2023seeing}, and lead to grave risks ranging from fake pornography to hoaxes, identity theft, and financial fraud \cite{tolosana2020deepfakes}. 
Therefore, it becomes extremely challenging to develop reliable synthetic content detection methods that can keep up with the latest generative models.

The scientific community has recently put a lot of effort on the development of automatic solutions to counter this problem \cite{mirsky2021creation,tolosana2020deepfakes}. More precisely, the discrimination of GAN-generated from real images using deep learning models has gained a lot of interest, although it has been found that such approaches struggle to generalize to other types of generative models \cite{cozzolino2018forensictransfer}. Synthetic Image Detection (SID) is typically performed based either on image-level \cite{wang2020cnn,chai2020makes} or frequency-level features \cite{frank2020leveraging,qian2020thinking,jeong2022bihpf}. Also, there are methods \cite{corvi2023intriguing,corvi2023detection} that analyse the traces left by generative models to provide useful insights on how to address the SID task.

\begin{figure}[t]
    \centering
    
    \subfloat[RINE (Ours)]{\includegraphics[width=.3\linewidth]{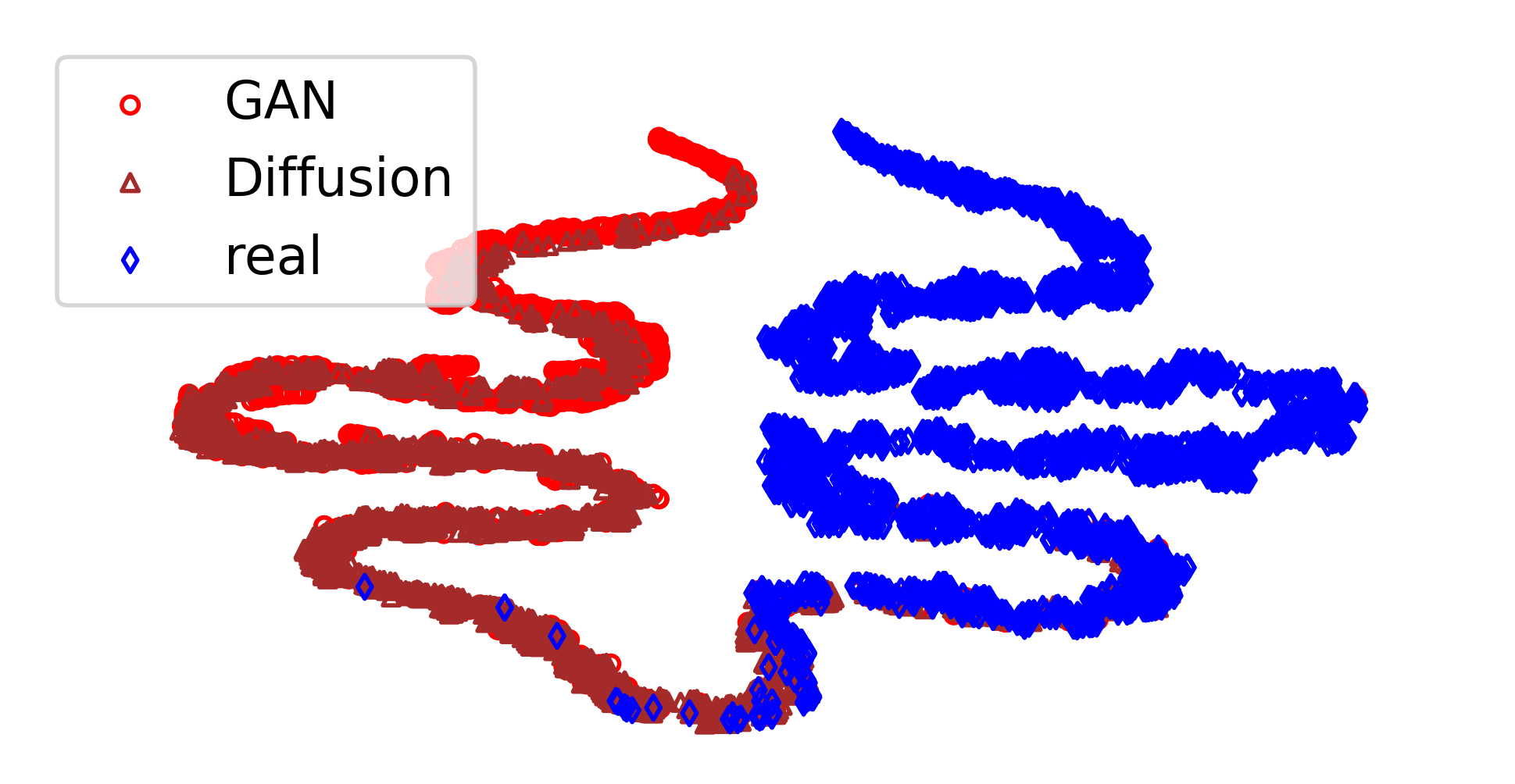}}
    \subfloat[UFD~\cite{ojha2023towards}]{\includegraphics[width=.3\linewidth]{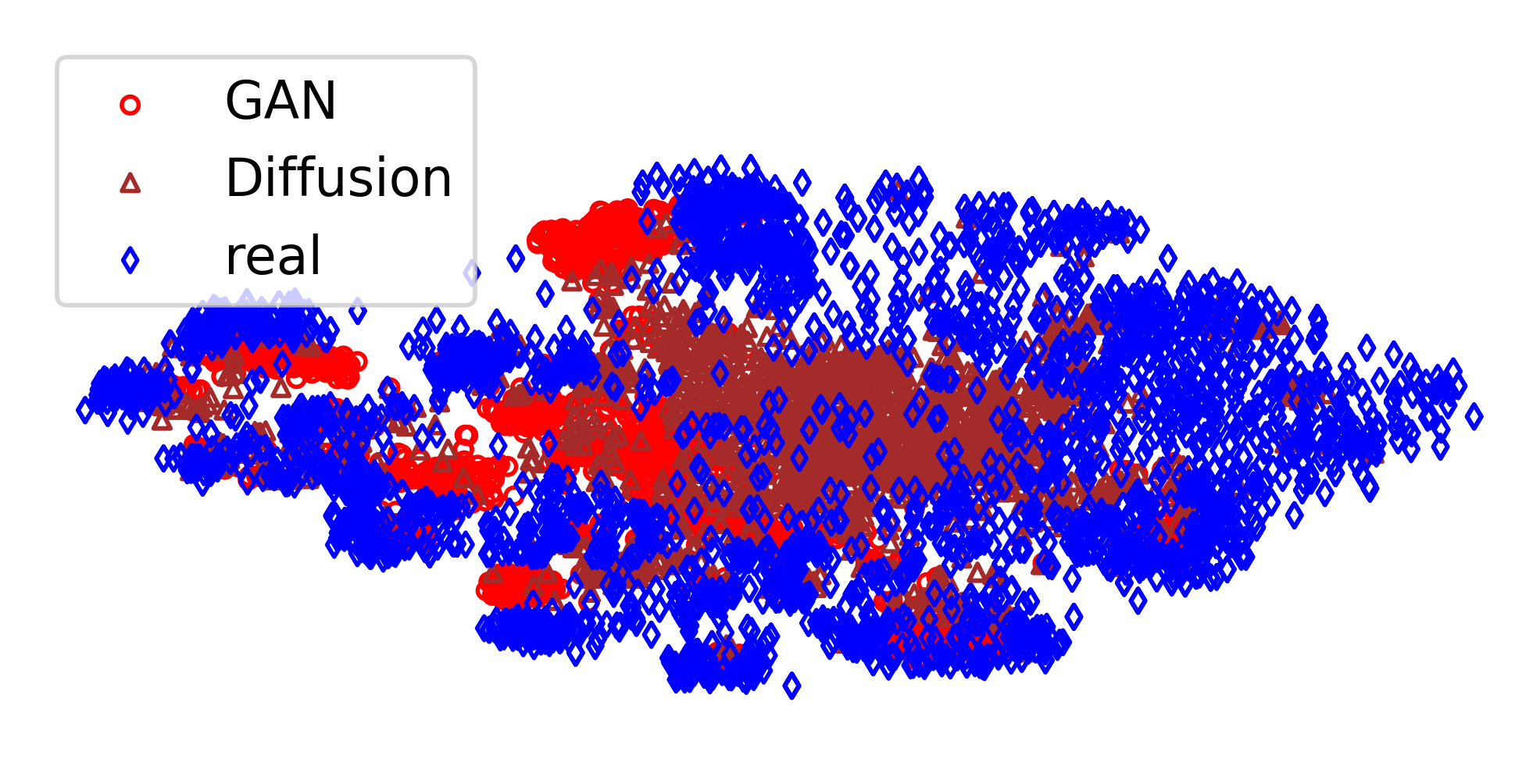}}
    \subfloat[Wang et al.~\cite{wang2020cnn}]{\includegraphics[width=.3\linewidth]{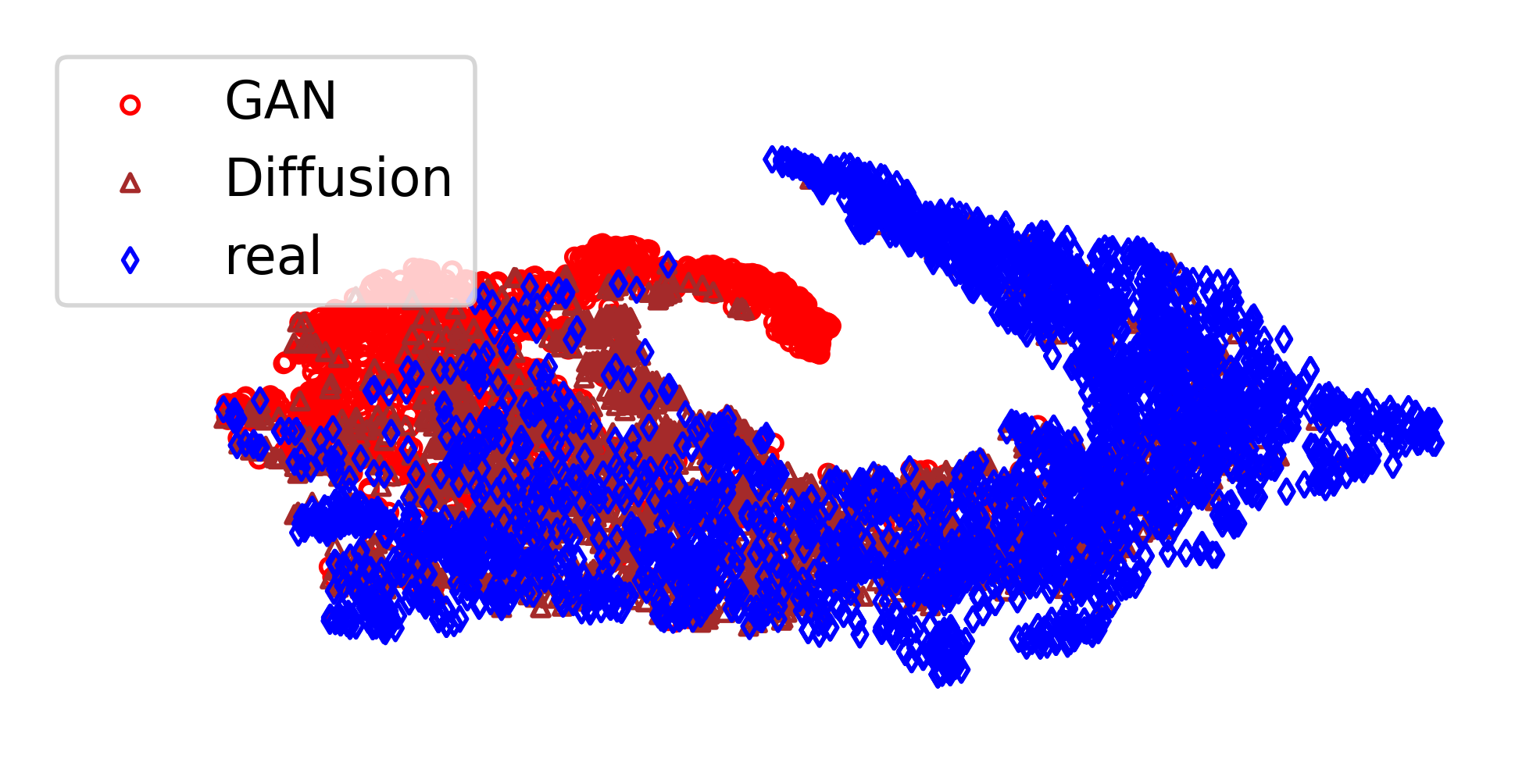}}
  
    \caption{Feature space visualization for unseen data with t-SNE.}
    \label{fig:tsne}
\end{figure}

Recently, the utilization of features extracted by foundation models, such as CLIP \cite{radford2021learning}, has been found to yield surprisingly high performance in the SID task, with minimal training requirements \cite{ojha2023towards}. However, these features are extracted from the last model layer, known to capture high-level semantics. Instead, low-level image features deriving from intermediate layers, which are known to be highly relevant for synthetic image detection \cite{bayar2018constrained,corvi2023detection}, have not been explored to date, despite their potential to further boost performance.

To address the SID task more effectively, we propose the RINE model by leveraging \underline{R}epresentations from \underline{In}termediate \underline{E}ncoder-blocks of CLIP. Specifically, we collect the image representations provided by intermediate Transformer blocks that carry low-level visual information and project them with learnable linear mappings to a forgery-aware vector space. Additionally, a Trainable Importance Estimator (TIE) module is used to incorporate the impact of each intermediate Transformer block in the final prediction. We train our models only on images generated by ProGAN (our experiments consider 1, 2, and 4 object classes for training) and evaluate it on 20 test datasets including images generated from GAN, Diffusion as well as other (deepfake, low-level vision, perceptual loss, DALL-E) generative models, surpassing by an average +10.6\% the state-of-the-art performance. \Cref{fig:tsne} provides a comparison between the feature space of RINE and that of state-of-the-art methods, for GAN, Diffussion and real images, showcasing the discriminating ability of the proposed representation. Also, it is noteworthy that (i) we reach this level of performance with only 6.3M learnable parameters, (ii) trained only for 1 epoch, which requires $\sim$8 minutes, and (iii) we surpass the state-of-the-art by +9.4\%, when training only on a single ProGAN object class, while the second-to-best model uses 20.

\section{Related Work}
\label{sec:related_work}
Synthetic Image Detection (SID) has recently emerged as a special but increasingly interesting field of DeepFake\footnote{Often used as an umbrella term for all kinds of synthetic or manipulated content} or synthetic media detection \cite{tolosana2020deepfakes, mirsky2021creation}. Early studies investigated fingerprints left by GAN generators, similarly to the fingerprints left by cameras \cite{lukas2006digital}, letting them not only to distinguish between real and synthetic images but also to attribute synthetic images to specific GAN generators \cite{marra2019gans,yu2019attributing}. Other studies focused on GAN-based image-to-image translation detection \cite{marra2018detection} and face manipulation detection \cite{rossler2019faceforensics++}, but still the considered evaluation sets contained images generated from the same generators as the ones used for training, neglecting to assess generalizability. Such approaches have been shown to struggle with detecting images coming from unseen generators \cite{cozzolino2018forensictransfer}.

Subsequent works focused on alleviating this issue through patch-based predictions~\cite{chai2020makes}, auto-encoder based architectures \cite{cozzolino2018forensictransfer} and co-occurrence matrices computed on the RGB image channels \cite{nataraj2019detecting}. Next, Wang et al. \cite{wang2020cnn} proposed a simple yet effective approach, based on appropriate pre-processing and intense data augmentation, which is capable of generalizing well to many GAN generators by simply training on ProGAN images \cite{karras2018progressive}. Additionally, traces of GAN generators in the frequency domain were exploited for effectively addressing the SID task \cite{zhang2019detecting,frank2020leveraging}, while the use of frequency-level perturbation maps was proposed in \cite{jeong2022frepgan} to force the detector to ignore domain-specific frequency-level artifacts. Image gradients estimated by a pre-trained CNN model have been shown to produce artifacts that can generalize well to unseen GAN generated images \cite{tan2023learning}. Removing the down-sampling operation from the first ResNet's layer, that potentially eliminates synthetic traces, together with intense augmentation has shown promising results \cite{corvi2023detection}. Another approach to improve generalization has been to select training images based on their perceptual quality; this was shown to be effective especially in cross-domain generalization settings \cite{dogoulis2023improving}.

Representations extracted by foundation models are surprisingly effective on SID, generalizing equally well on GAN and Diffusion model generated data \cite{ojha2023towards}. Motivated by this success, we hypothesize that further performance gains are possible by leveraging representations from intermediate layers, which carry low-level visual information, in addition to representations from the final layer that primarily carry high-level semantic information. 
Our approach is different from deep layer aggregation that was explored in several computer vision tasks \cite{yu2018deep}, as well as for deepfake detection \cite{jevnisek2022aggregating}, as we process the frozen layers' outputs of a foundation model, while deep layer aggregation is a training technique that concatenates layer outputs to produce predictions during the training of a CNN. 

Finally, the performance drop of synthetic image detectors in real-world conditions typically performed when uploading content to social media (e.g., cropping and compression) has been analysed in \cite{marra2018detection,corvi2023detection}. Our method is also effective in such conditions as demonstrated by the experimental results in \cref{subsec:robustness}.

\section{Methodology}
\label{sec:methodology}

\begin{figure}[t]
\centering
\includegraphics[width=\textwidth]{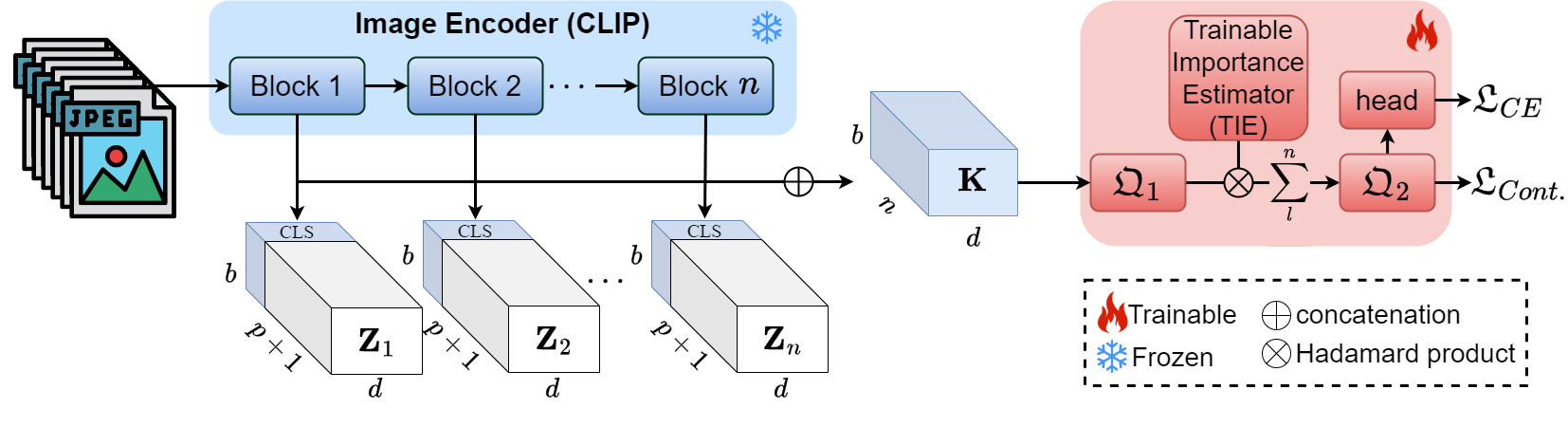}
\caption{
The RINE architecture. A batch of $b$ images is processed by CLIP's image encoder. The concatenation of the $n$ $d$-dimensional CLS tokens (one from each Transformer block) is first projected and then multiplied with the blocks' scores, estimated by the Trainable Importance Estimator (TIE) module. Summation across the second dimension results in one feature vector per image. Finally, after the second projection and the consequent classification head modules, two loss functions are computed. Binary cross-entropy $\mathfrak{L}_{CE}$ directly optimizes SID, while the contrastive loss $\mathfrak{L}_{Cont.}$ assists the training by forming a dense feature vector cluster per class.
}\label{fig:architecture}
\end{figure}

\subsection{Representations from Intermediate Encoder-Blocks}
\Cref{fig:architecture} illustrates the RINE architecture. 
Let us consider a batch of $b$ input images $\mathbf{X}\in\mathbb{R}^{b\times 3\times w\times h}$, with $w$ width, and $h$ height, and targets $\mathbf{Y}=\{y_i\}_{i=1}^b\in\mathbb{R}^b$, with $y_i\in\{0,1\}$.
The images are reshaped into a sequence of $p$ flattened patches $\mathbf{X}_p\in\mathbb{R}^{b\times p\times (P^2\cdot 3)}$, where $P$ denotes patch side length and $p=w\cdot h/P^2$.
Then, $\mathbf{X}_p$ is linearly projected to $d$ dimensions, the learnable $d$-dimensional CLS token is concatenated to the projected sequence, and positional embeddings are added to all $p+1$ tokens in order to construct the image encoder's input, $\mathbf{Z}_0\in\mathbb{R}^{b\times (p+1)\times d}$. CLIP's image encoder~\cite{radford2021learning}, being a Vision Transformer (ViT)~\cite{dosovitskiy2020image}, consists of $n$ successive Transformer encoder blocks~\cite{vaswani2017attention}, each processing the output of the previous block as shown in \cref{eq:vit}:
\begin{equation}
\label{eq:vit}
\begin{split}
    \tilde{\mathbf{Z}_l}&=\text{MSA}(\text{LN}(\mathbf{Z}_{l-1}))+\mathbf{Z}_{l-1}\\
    \mathbf{Z}_l&=\text{MLP}(\text{LN}(\tilde{\mathbf{Z}_l}))+\tilde{\mathbf{Z}_l}
\end{split}
\end{equation}
where $l$=1,\dots,n denotes the block's index, $\mathbf{Z}_l\in\mathbb{R}^{b\times (p+1)\times d}$ denotes the $l$th block's output, MSA denotes Multi-head Softmax Attention~\cite{vaswani2017attention}, LN denotes Layer Normalization~\cite{ba2016layer}, and MLP denotes a Multi-Layer Perceptron with two GELU~\cite{hendrycks2016gaussian} activated linear layers of $4\cdot d$ and $d$ number of units, respectively.

Then, we define the Representations from Intermediate Encoder-blocks (RINE) $\mathbf{K}$, as the concatenation of CLS tokens from the corresponding $n$ blocks:
\begin{equation}
    \mathbf{K}=\oplus\big\{\mathbf{Z}_l^{\text{[0]}}\big\}_{l=1}^n\in\mathbb{R}^{b\times n\times d}
\end{equation}
where $\oplus$ denotes concatenation, and $\mathbf{Z}_l^{\text{[0]}}\in\mathbb{R}^{b\times 1\times d}$ denotes the CLS token from the output of block $l$. We keep CLIP, which produces these features, frozen during training and use $\mathbf{K}$ to construct discriminative features for the SID task.

\subsection{Trainable Modules}
The representations $\mathbf{K}$ are processed by a projection network ($\mathfrak{Q}_1$ in \cref{fig:architecture}):
\begin{equation}
\label{eq:projection}
    \mathbf{K}_m=\text{ReLU}(\mathbf{K}_{m-1}\mathbf{W}_{m}+\mathbf{b}_{m})\in\mathbb{R}^{b\times n\times d'}
\end{equation}
where $m$=1,\dots,q denotes the index of the network's layer (with $\mathbf{K}_0=\mathbf{K}$), $\mathbf{W}_m\in\mathbb{R}^{d'\times d'}$ (except $\mathbf{W}_1\in\mathbb{R}^{d\times d'}$) and $\mathbf{b}_m\in\mathbb{R}^{d'}$ define the linear mapping, and ReLU denotes the Rectified Linear Unit \cite{agarap2018deep} activation function. After each layer, dropout \cite{srivastava2014dropout} with rate 0.5 is applied.

Additionally, each of the $d'$ learned features can be more relevant for the SID task at various processing stages, thus we employ a Trainable Importance Estimator (TIE) module to adjust their impact to the final decision. More precisely, we consider a randomly-initialized learnable variable $\mathbf{A}=\{\alpha_{lk}\}\in\mathbb{R}^{n\times d'}$ the elements of which will estimate the importance of feature $k$ at processing stage (i.e., Transformer block) $l$. This is used to construct one feature vector per image $i$, through weighted averages of features $k$ across processing stages $l$:
\begin{equation}\label{eq:tie}
    \tilde{\mathbf{K}}^{(ik)}=\sum_l^n\mathcal{S}(\mathbf{A})^{(lk)}\cdot\mathbf{K}_q^{(ilk)}
\end{equation}
where $^{(\cdot)}$ denotes tensor indices, $\mathbf{K}_q=\{\mathbf{K}_q^{(ilk)}\}\in\mathbb{R}^{b\times n\times d'}$ is the output of projection $\mathfrak{Q}_1$, and $\mathcal{S}$ denotes the Softmax activation function acting across the first dimension of $\mathbf{A}$.

Finally, a second projection network ($\mathfrak{Q}_2$ in \cref{fig:architecture}) with the same architecture as the first one (cf. \cref{eq:projection}) takes as input $\tilde{\mathbf{K}}\in\mathbb{R}^{b\times d'}$ and outputs $\tilde{\mathbf{K}}_q\in\mathbb{R}^{b\times d'}$, which is consequently processed by the classification head that predicts the final output $\hat{y}_i$ (probability to be fake). The classification head consists of two $d'\times d'$ ReLU-activated dense layers and one $d'\times 1$ dense layer which produces the logits.

\subsection{Objective Function}
We consider the combination of two objective functions to optimize the parameters of the proposed model. The first is the binary cross-entropy loss~\cite{good1952rational}, which measures the classification error and directly optimizes the SID objective:
\[\mathfrak{L}_{CE}=-\sum_{i=1}^by_i\text{log}\hat{y}_i+(1-y_i)\cdot\text{log}(1-\hat{y}_i)\]
The second is the Supervised Contrastive Learning loss (SupContrast)~\cite{khosla2020supervised}, which is considered in order to assist the training process by bringing closer the feature vectors inside $\tilde{\mathbf{K}}_q$ that share targets and move apart the rest, defined as:
\[\mathfrak{L}_{Cont.}=-\sum_{i=1}^b\frac{1}{G(i)}\sum_{g\in G(i)}\text{log}\frac{\text{exp}(\mathbf{z}_i\cdot \mathbf{z}_g/\tau)}{\sum_{a\in A(i)}\text{exp}(\mathbf{z}_i\cdot \mathbf{z}_a/\tau)}\]
where $A(i)=\{1,\dots,i-1,i+1,\dots,b\}$, $G(i)=\{g\in A(i):y_g=y_i\}$, $\mathbf{z}_j=\{\tilde{\mathbf{K}}_q^{(jk)}\}_{k=1}^{d'}\in\mathbb{R}^{d'}$, the $\cdot$ denotes dot product, and $\tau$ is a temperature parameter.
We combine the two loss functions by a tunable factor $\xi$, as shown in \cref{eq:loss}:
\begin{equation}
\label{eq:loss}
\mathfrak{L}=\mathfrak{L}_{CE}+\xi\cdot\mathfrak{L}_{Cont.}
\end{equation}

\section{Experimental Setup}
\label{sec:experimental_setup}
\subsection{Datasets}
\label{subsec:datasets}
Following the training protocol of related work~\cite{ojha2023towards,wang2020cnn,tan2023learning}, we use for training ProGAN~\cite{karras2018progressive} generated images and the corresponding real images of the provided dataset. Similarly with previous works \cite{tan2023learning,jeong2022frepgan}, we consider three settings with ProGAN-generated training data from 1 (horse), 2 (chair, horse) and 4 (car, cat, chair, horse) object classes. For testing, this is the first study to consider 20 datasets, with generated and real images, combining the evaluation datasets used in \cite{ojha2023towards,wang2020cnn,tan2023learning}. Specifically, the evaluation sets are from ProGAN~\cite{karras2018progressive}, StyleGAN~\cite{karras2019style}, StyleGAN2~\cite{karras2020analyzing}, BigGAN~\cite{brock2018large}, CycleGAN~\cite{zhu2017unpaired}, StarGAN~\cite{choi2018stargan}, GauGAN~\cite{park2019semantic}, DeepFake~\cite{rossler2019faceforensics++}, SITD~\cite{chen2018learning}, SAN~\cite{dai2019second}, CRN~\cite{chen2017photographic}, IMLE~\cite{li2019diverse}, Guided~\cite{dhariwal2021diffusion}, LDM~\cite{rombach2022high} (3 variants), Glide~\cite{nichol2021glide} (3 variants), and DALL-E~\cite{ramesh2021zero}.

\subsection{Implementation Details}
\label{subsec:implementation}
The training of RINE is conducted with batch size 128 and learning rate 1e-3 for only 1 epoch using the Adam optimizer. To assess the existence of potential benefit from further training we also conduct a set of experiments with 3, 5, 10, and 15 epochs (cf. \cref{subsec:training_duration}). For each of the 3 training settings (1-class, 2-class, 4-class), we consider a hyperparameter grid to obtain the best performance, namely $\xi\in\{0.1, 0.2, 0.4, 0.8\}$, $q\in\{1, 2, 4\}$, and $d'\in~\{128, 256, 512, 1024\}$. Additionally, we consider two CLIP variants, namely ViT-B/32 and L/14, for the extraction of representations. The models' hyperparameters that result in the best performance (presented in \cref{sec:results}) are illustrated in \cref{tab:hyperparams}, along with the corresponding number of trainable parameters, and required training duration. For the training images, following the best practice in related work, we  
apply Gaussian blurring and $\mathtt{JPEG}$ compression with probability 0.5, then random cropping to 224$\times$224, and finally random horizontal flip with probability 0.5. The validation and test images are 
only center-cropped at 224$\times$224. Resizing is omitted both during training and testing as it is known to eliminate synthetic traces \cite{corvi2023detection}. All experiments are conducted using one NVIDIA GeForce RTX 3090 Ti GPU.

\begin{table}[h]
\caption{Hyperparameter configuration of the best 1-, 2-, and 4-class models. The number of learnable parameters per model (in millions), and the training time (in minutes) is also reported.}
    \label{tab:hyperparams}
    \centering
    \begin{tabular}{cccccccc}
    \toprule
        \# classes&backbone&$\xi$&$q$&$d$&$d'$&\# params. &training time\\
        \midrule
         1&ViT-L/14&0.1&4&1024&1024&10.52M&$\sim$2 min.\\
         2&ViT-L/14&0.2&4&1024&128&0.28M&$\sim$4 min.\\
         4&ViT-L/14&0.2&2&1024&1024&6.32M&$\sim$8 min.\\
    \bottomrule
    \end{tabular}
    
\end{table}

\subsection{State-of-the-Art Methods}
The state-of-the-art SID methods we consider in our comparison include the following:
\begin{enumerate}
    \item Wang~\cite{wang2020cnn}: A standard ResNet-50 \cite{he2016deep} architecture pre-trained on ImageNet is fine-tuned on SID with appropriate selection of pre- and post-processing, as well as data augmentations.
    \item Patch-Forensics~\cite{chai2020makes}: A truncated (at block 2) Xception~\cite{chollet2017xception} pre-trained on ImageNet is fine-tuned on SID for each image patch, finally aggregating the individual patch-level predictions.
    \item FrePGAN~\cite{jeong2022frepgan}:  Frequency-level perturbation maps are generated by an adversarially trained network making fake images hard to distinguish from real ones, then a standard ResNet-50 \cite{he2016deep}  is trained on the SID task.
    \item LGrad~\cite{tan2023learning}: Image gradients, computed using a pre-trained deep network, are processed by a standard ResNet-50 \cite{he2016deep}  pre-trained on ImageNet, which is fine-tuned on the SID task.
    \item DMID~\cite{corvi2023detection}: A ResNet-50 \cite{he2016deep} without down-sampling in the first layer is trained on the SID task using intense augmentation.
    \item Universal Fake Detector (UFD)~\cite{ojha2023towards}: Simple linear probing on top of (last layer) features extracted from CLIP's ViT-L/14 \cite{radford2021learning} image encoder.
\end{enumerate}

For all methods except FrePGAN, we compute performance metrics on the evaluation datasets presented in \cref{subsec:datasets}, using the publicly available checkpoints provided by their official repositories. For FrePGAN, that does not provide publicly available code and models, we present the scores that are reported in their paper \cite{jeong2022frepgan}.

\subsection{Evaluation Protocol}
We evaluate the proposed architecture with accuracy (ACC) and average precision (AP) metrics on each test dataset, following previous works for comparability purposes. For the calculation of accuracy no calibration is conducted, we consider 0.5 as threshold to all methods. Best models are identified by the maximum sum ACC+AP. We also report the average (AVG) metric values across the test datasets to obtain summary evaluations.

\section{Results}
\label{sec:results}

\subsection{Comparative Analysis}
In \cref{tab:comparative_accuracy} and \cref{tab:comparative_ap}, we present the performance scores (ACC \& AP respectively) of our method versus the competing ones. Our 1-class model outperforms all state-of-the-art methods irrespective of training class number. 
On average, we surpass the state-of-the-art by +9.4\% ACC \& +4.3\% AP with the 1-class model, by +6.8\% ACC \& +4.4\% AP with the 2-class model, and by +10.6\% ACC \& +4.5\% AP with the 4-class model. In terms of ACC, we obtain the best score in 14 out of 20 test datasets, and  simultaneously the first and second best performance in 10 of them. In terms of AP, we obtain the best score in 15 out of 20 test datasets, and simultaneously the first and second best performance in 14 of them. The biggest performance gain is on the SAN dataset \cite{dai2019second} (+16.9\% ACC). Also, considering more training classes does not reduce generalization (cf. Appendix).

\begin{table}[t]
\caption{Accuracy (ACC) scores of baselines and our model across 20 test datasets. The second column (\# cl.) presents the number of used training classes. Best performance is denoted with \textbf{bold} and second to best with \underline{underline}. Our method yields +10.6\% average accuracy compared to the state-of-the-art.}
    \label{tab:comparative_accuracy}
    \resizebox{\textwidth}{!}{
    \begin{tabular}{lcccccccccccccccccccccccc}
    \toprule
        &&\multicolumn{8}{c}{Generative Adversarial Networks}&\multicolumn{2}{c}{Low level vision}&&\multicolumn{2}{c}{Perceptual loss}&&\multicolumn{3}{c}{Latent Diffusion}&&\multicolumn{3}{c}{Glide}\\
        \cmidrule{3-9}\cmidrule{11-12}\cmidrule{14-15}\cmidrule{17-19}\cmidrule{21-23}\\
        && Pro- & Style- & Style- & Big- & Cycle- & Star- & Gau- & Deep- &&&&&&&200&200&100&&100&50&100&&AVG\\
        method&\# cl.&GAN&GAN&GAN2&GAN&GAN&GAN&GAN&fake&SITD&SAN&&CRN&IMLE&Guided&steps& CFG&steps&&27&27&10&DALL-E&\\
        \midrule

        Wang~\cite{wang2020cnn} (prob. 0.5)&20&\textbf{100.0}&66.8&64.4&59.0&80.7&80.9&79.2&51.3&55.8&50.0&&85.6&92.3&52.1&51.1&51.4&51.3&&53.3&55.6&54.2&52.5&64.4\\
        
        Wang~\cite{wang2020cnn} (prob. 0.1)&20&\textbf{100.0}&84.3&82.8&70.2&85.2&91.7&78.9&53.0&63.1&50.0&&90.4&90.3&60.4&53.8&55.2&55.1&&60.3&62.7&61.0&56.0&70.2\\

        Patch-Forensics~\cite{chai2020makes}&$\dagger$&66.2&58.8&52.7&52.1&50.2&96.9&50.1&58.0&54.4&50.0&&52.9&52.3&50.5&51.9&53.8&52.0&&51.8&52.1&51.4&57.2&55.8\\
        FrePGAN~\cite{jeong2022frepgan}&1&95.5&80.6&77.4&63.5&59.4&99.6&53.0&70.4&-*&-&&-&-&-&-&-&-&&-&-&-&-&-\\
        FrePGAN~\cite{jeong2022frepgan}&2&99.0&80.8&72.2&66.0&69.1&98.5&53.1&62.2&-&-&&-&-&-&-&-&-&&-&-&-&-&-\\
        FrePGAN~\cite{jeong2022frepgan}&4&99.0&80.7&84.1&69.2&71.1&\textbf{99.9}&60.3&70.9&-&-&&-&-&-&-&-&-&&-&-&-&-&-\\
        LGrad~\cite{tan2023learning}&1&99.4&\underline{96.1}&94.0&79.6&84.6&99.5&71.1&63.4&50.0&44.5&&52.0&52.0&67.4&90.5&\underline{93.2}&90.6&&80.2&85.2&83.5&89.5&78.3\\
        LGrad~\cite{tan2023learning}&2&99.8&94.5&92.1&82.5&85.5&\underline{99.8}&73.7&61.5&46.9&45.7&&52.0&52.1&72.1&91.1&93.0&91.2&&87.1&90.5&89.4&88.7&79.4\\
        LGrad~\cite{tan2023learning}&4&\underline{99.9}&94.8&\textbf{96.1}&83.0&85.1&99.6&72.5&56.4&47.8&41.1&&50.6&50.7&74.2&94.2&\textbf{95.9}&95.0&&\underline{87.2}&\underline{90.8}&\underline{89.8}&88.4&79.7\\

        DMID~\cite{corvi2023detection}&20&\textbf{100.0}&\textbf{99.4}&92.9&96.9&92.0&99.5&94.8&54.1&\underline{90.6}**&55.5&&\textbf{100.0}&\textbf{100.0}&53.9&58.0&61.1&57.5&&56.9&59.6&58.8&71.7&77.6\\
        UFD~\cite{ojha2023towards}&20&99.8&79.9&70.9&95.1&98.3&95.7&99.5&71.7&71.4&51.4&&57.5&70.0&70.2&94.4&74.0&95.0&&78.5&79.0&77.9&87.3&80.9\\
        \midrule        
        
        &1&99.8&88.7&86.9&\underline{99.1}&\textbf{99.4}&98.8&99.7&\textbf{82.7}&84.7&\textbf{72.4}&&\underline{93.4}&\underline{96.9}&\textbf{77.9}&\underline{96.9}&83.5&97.0&&83.8&87.4&85.4&91.9&\underline{90.3}\\ 
        RINE (Ours) &2&99.8&84.9&76.7&98.3&\textbf{99.4}&99.6&\textbf{99.9}&66.7&\textbf{91.9}&67.8&&83.5&96.8&69.6&96.8&80.0&\underline{97.3}&&83.6&86.0&84.1&\underline{92.3}&87.7\\
        &4&\textbf{100.0}&88.9&\underline{94.5}&\textbf{99.6}&\underline{99.3}&99.5&\underline{99.8}&\underline{80.6}&\underline{90.6}&\underline{68.3}&&89.2&90.6&\underline{76.1}&\textbf{98.3}&88.2&\textbf{98.6}&&\textbf{88.9}&\textbf{92.6}&\textbf{90.7}&\textbf{95.0}&\textbf{91.5}\\
        \bottomrule
    \end{tabular}
    }
    \tiny * Hyphens denote scores that are neither reported in the corresponding paper nor the code and models are available in order to compute them.\\
    \tiny ** We applied cropping at 2000x1000 on SITD \cite{chen2018learning} for DMID \cite{corvi2023detection} due to GPU memory limitations.\\
    \tiny $\dagger$ Patch-Forensics has been trained on ProGAN data but not on the same dataset as the rest models. For more details please refer to \cite{chai2020makes}.
    
\end{table}

\begin{table}[]
\caption{Average precision (AP) scores of baselines and our model across 20 test datasets. The second column (\# cl.) presents the number of used training classes. Best performance is denoted with \textbf{bold} and second to best with \underline{underline}. Our method yields +4.5\% mean average precision (mAP) compared to the state-of-the-art.}
    \label{tab:comparative_ap}
    \resizebox{\textwidth}{!}{\begin{tabular}{lcccccccccccccccccccccccc}
    \toprule
        &&\multicolumn{8}{c}{Generative Adversarial Networks}&\multicolumn{2}{c}{Low level vision}&&\multicolumn{2}{c}{Perceptual loss}&&\multicolumn{3}{c}{Latent Diffusion}&&\multicolumn{3}{c}{Glide}\\
        \cmidrule{3-9}\cmidrule{11-12}\cmidrule{14-15}\cmidrule{17-19}\cmidrule{21-23}\\
        && Pro- & Style- & Style- & Big- & Cycle- & Star- & Gau- & Deep- &&&&&&&200&200&100&&100&50&100&&AVG\\
        method&\# cl.&GAN&GAN&GAN2&GAN&GAN&GAN&GAN&fake&SITD&SAN&&CRN&IMLE&Guided&steps& CFG&steps&&27&27&10&DALL-E&\\
        \midrule
        Wang~\cite{wang2020cnn} (prob. 0.5)&20&\textbf{100.0}&98.0&97.8&88.2&96.8&95.4&98.1&64.8&82.2&56.0&&99.4&99.7&69.9&65.9&66.7&66.0&&72.0&76.5&73.2&66.3&81.7\\
        
        Wang~\cite{wang2020cnn} (prob. 0.1)&20&\textbf{100.0}&99.5&99.0&84.5&93.5&98.2&89.5&87.0&68.1&53.0&&99.5&99.5&73.2&71.2&73.0&72.5&&80.5&84.6&82.1&71.3&84.0\\

        Patch-Forensics~\cite{chai2020makes}&$\dagger$&94.6&79.3&77.6&83.3&74.7&99.5&83.2&71.3&91.6&39.7&&\underline{99.9}&98.9&58.7&68.9&73.7&68.7&&50.6&52.8&48.4&66.9&74.1\\
        FrePGAN~\cite{jeong2022frepgan}&1&99.4&90.6&93.0&60.5&59.9&\textbf{100.0}&49.1&81.5&-*&-&&-&-&-&-&-&-&&-&-&-&-&-\\
        FrePGAN~\cite{jeong2022frepgan}&2&\underline{99.9}&92.0&94.0&61.8&70.3&\textbf{100.0}&51.0&80.6&-&-&&-&-&-&-&-&-&&-&-&-&-&-\\
        FrePGAN~\cite{jeong2022frepgan}&4&\underline{99.9}&89.6&98.6&71.1&74.4&\textbf{100.0}&71.7&91.9&-&-&&-&-&-&-&-&-&&-&-&-&-&-\\
        LGrad~\cite{tan2023learning}&1&\underline{99.9}&99.6&99.5&88.9&94.4&\textbf{100.0}&82.0&79.7&44.1&45.7&&82.3&82.5&71.1&97.3&98.0&97.2&&90.1&93.6&92.0&96.9&86.7\\
        LGrad~\cite{tan2023learning}&2&\textbf{100.0}&99.6&99.6&92.6&94.7&\underline{99.9}&83.2&71.6&42.4&45.3&&66.1&80.9&75.6&97.2&98.1&97.2&&94.6&96.5&95.8&96.5&86.4\\
        LGrad~\cite{tan2023learning}&4&\textbf{100.0}&\underline{99.8}&\underline{99.9}&90.8&94.0&\textbf{100.0}&79.5&72.4&39.4&42.2&&63.9&69.7&79.5&99.1&\textbf{99.1}&99.2&&93.3&95.2&95.0&97.3&85.5\\

        DMID~\cite{corvi2023detection}&20&\textbf{100.}0&\textbf{100.0}&\textbf{100.0}&\underline{99.8}&98.6&\textbf{100.0}&\underline{99.8}&94.7&\textbf{99.8}**&87.7&&\textbf{100.0}&\textbf{100.0}&73.0&86.8&89.4&87.3&&86.5&89.9&89.0&96.1&93.9\\
        UFD~\cite{ojha2023towards}&20&\textbf{100.0}&97.3&97.5&99.3&\underline{99.8}&99.4&\textbf{100.0}&84.4&89.9&62.6&&94.5&98.3&89.5&99.3&92.5&\underline{99.3}&&95.3&95.6&95.0&97.5&94.3\\
        \midrule
        
        &1&\textbf{100.0}&99.1&99.7&\textbf{99.9}&\textbf{100.0}&\textbf{100.0}&\textbf{100.0}&\underline{97.4}&95.8&91.9&&98.5&\underline{99.9}&\underline{95.7}&\underline{99.8}&98.0&\textbf{99.9}&&\textbf{98.9}&\textbf{99.3}&\textbf{99.1}&\underline{99.3}&98.6\\
        RINE (Ours) &2&\textbf{100.0}&99.5&99.6&\textbf{99.9}&\textbf{100.0}&\textbf{100.0}&\textbf{100.0}&96.4&\underline{97.5}&\underline{93.1}&&98.2&99.8&\underline{95.7}&\textbf{99.9}&98.0&\textbf{99.9}&&\textbf{98.9}&\underline{99.0}&98.8&\textbf{99.6}&\underline{98.7}\\
        
        &4&\textbf{100.0}&99.4&\textbf{100.0}&\textbf{99.9}&\textbf{100.0}&\textbf{100.0}&\textbf{100.0}&\textbf{97.9}&97.2&\textbf{94.9}&&97.3&99.7&\textbf{96.4}&\underline{99.8}&\underline{98.3}&\textbf{99.9}&&\underline{98.8}&\textbf{99.3}&\underline{98.9}&\underline{99.3}&\textbf{98.8}\\
        \bottomrule
    \end{tabular}}
    \tiny * Hyphens denote scores that are neither reported in the corresponding paper nor the code and models are available in order to compute them.\\
    \tiny ** We applied cropping at 2000x1000 on SITD \cite{chen2018learning} for DMID \cite{corvi2023detection} due to GPU memory limitations.\\
    \tiny $\dagger$ Patch-Forensics has been trained on ProGAN data but not on the same dataset as the rest models. For more details please refer to \cite{chai2020makes}.
    
\end{table}

\subsection{Ablations}\label{subsec:ablations}
In \cref{tab:ablations}, we present an ablation study, where we remove three of RINE's main components, namely the intermediate representations, the TIE, and the contrastive loss, one-by-one. To be more precise, ``w/o intermediate'' means that we use only the last layer's features (equivalent to the SotA method \cite{ojha2023towards}). We measure the RINE's performance on 20 test datasets, after removing each of the three components, and present the average ACC and AP for the 1-, 2-, and 4-class models, as well as the average across the three models. In addition, we present generalizability results by averaging across the non-GAN generators. The results demonstrate the positive impact of all proposed components.  
Also, ablating intermediate representations yields the biggest performance loss reducing the metrics to the previous SotA levels (i.e., \cite{ojha2023towards}). In the Appendix the interested reader can find ablation of the fusion mechanism and backbone.

\begin{table}[t]
\caption{Ablation analysis compares the full architecture with it after removing the contrastive loss, the TIE module, and the intermediate representations.}
    \label{tab:ablations}
    \centering
    \begin{tabular}{llllllllllll}
    \toprule
        &\multicolumn{2}{c}{1-class}&&\multicolumn{2}{c}{2-class}&&\multicolumn{2}{c}{4-class}&&\multicolumn{2}{c}{AVG}\\
        &ACC&AP&&ACC&AP&&ACC&AP&&ACC&AP\\
        \cmidrule{2-3}\cmidrule{5-6}\cmidrule{8-9}\cmidrule{11-12}
        \midrule
        \multicolumn{12}{c}{all generators}\\
        \midrule
        w/o contr. loss & 87.3&98.3&&\textbf{87.9}&98.5&&90.0&98.8&&88.4&98.5\\
        w/o TIE & 85.0&97.8&&86.4&98.5&&90.5&98.8&&87.3&98.3\\
        w/o intermediate &78.9&93.1&&81.1&94.7&&82.5&94.8&&80.8&94.2\\
        \midrule
        full & \textbf{90.3}&\textbf{98.6} &&87.7&\textbf{98.7}&&\textbf{91.5}&\textbf{98.8}&&\textbf{89.8}&\textbf{98.7}\\
        \midrule
        \multicolumn{12}{c}{non-GAN generators}\\
        \midrule
        w/o contr. loss & 83.4&97.4&&\textbf{84.6}&97.7&&86.3&98.2&&84.7&97.8\\
        w/o TIE & 80.0&96.6&&82.0&97.8&&87.0&98.2&&83.0&97.5\\
        w/o intermediate &73.4&90.0&&76.3&92.4&&77.0&92.4&&75.5&91.6\\
        \midrule
        full & \textbf{87.2}&\textbf{98.0} &&84.3&\textbf{98.1}&&\textbf{88.3}&\textbf{98.3}&&\textbf{86.6}&\textbf{98.1}\\
    \bottomrule
    \end{tabular}
    
\end{table}

\subsection{Hyperparameter Analysis}
\begin{figure}[t]
    \centering
    \subfloat[]{\includegraphics[width=.33\linewidth]{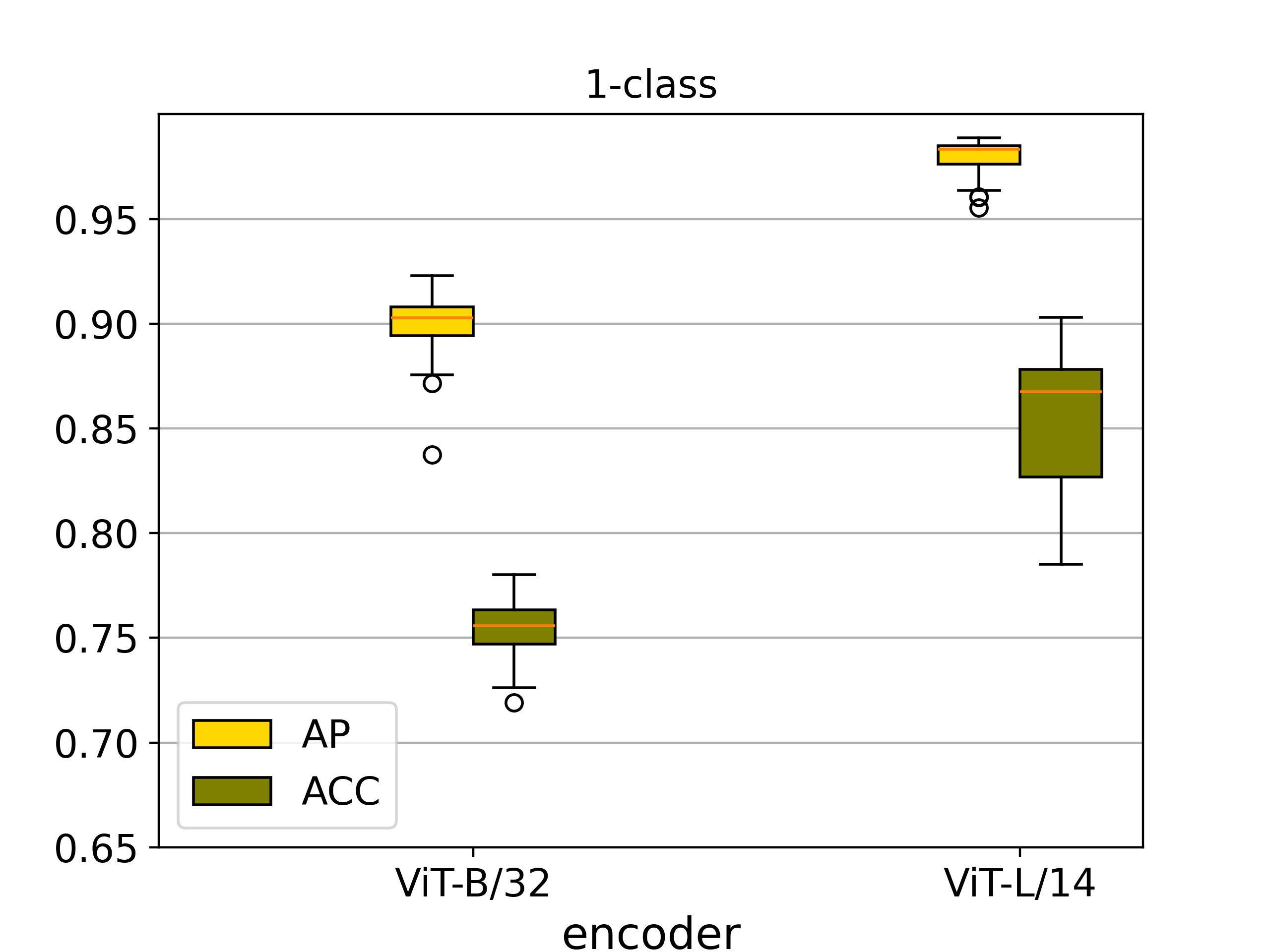}}
    \subfloat[]{\includegraphics[width=.33\linewidth]{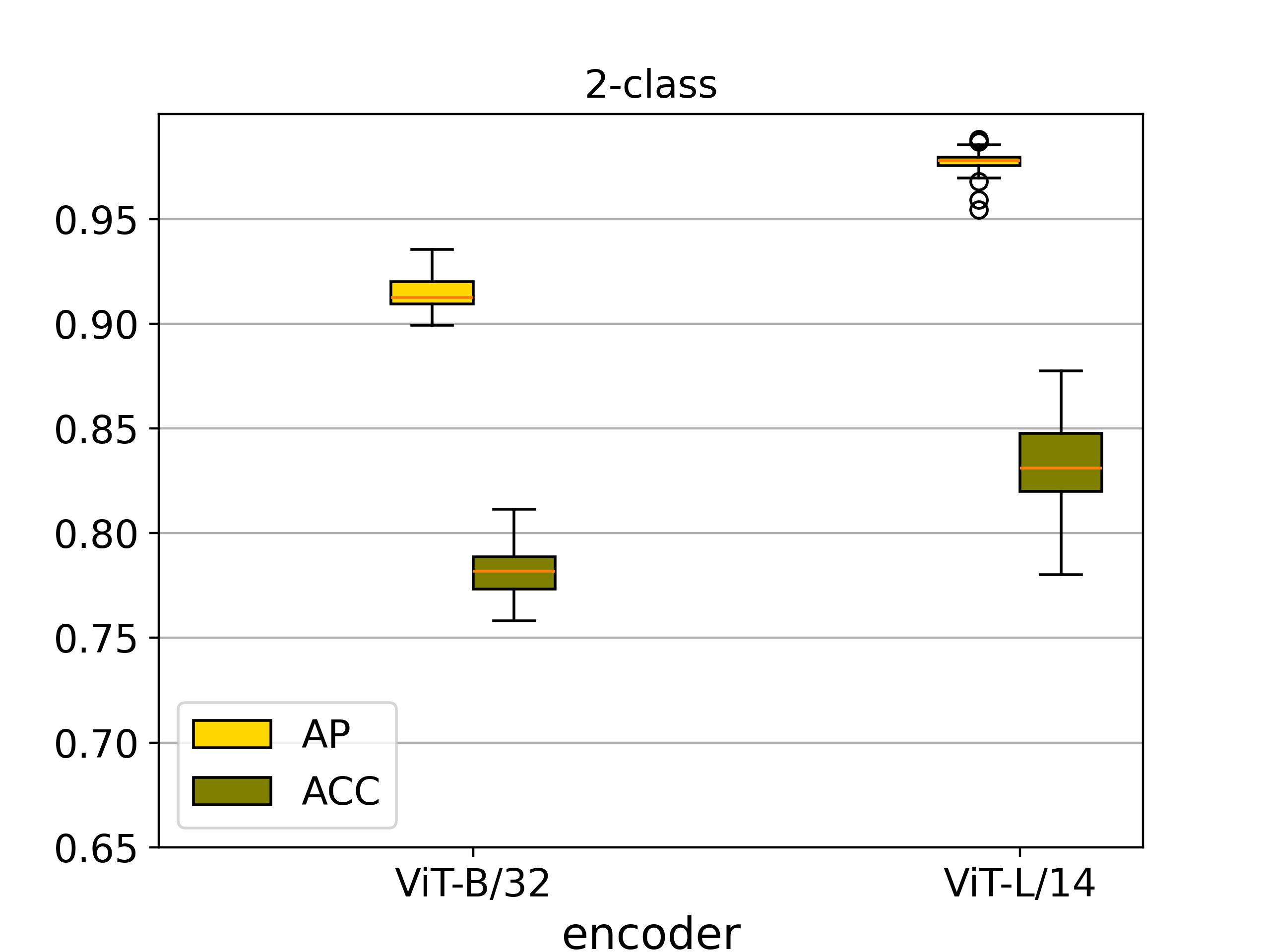}}
    \subfloat[]{\includegraphics[width=.33\linewidth]{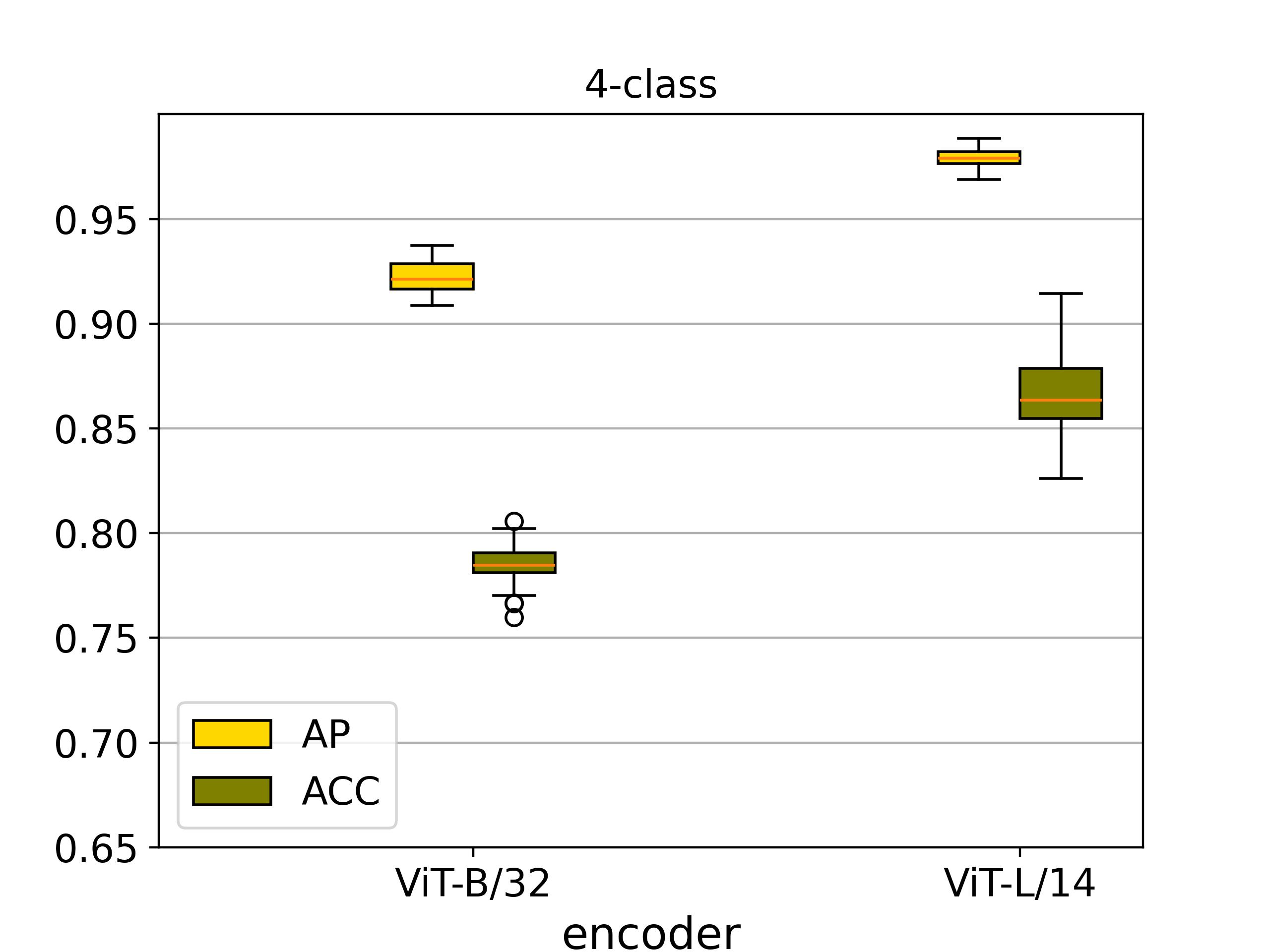}}

    \subfloat[]{\includegraphics[width=.33\linewidth]{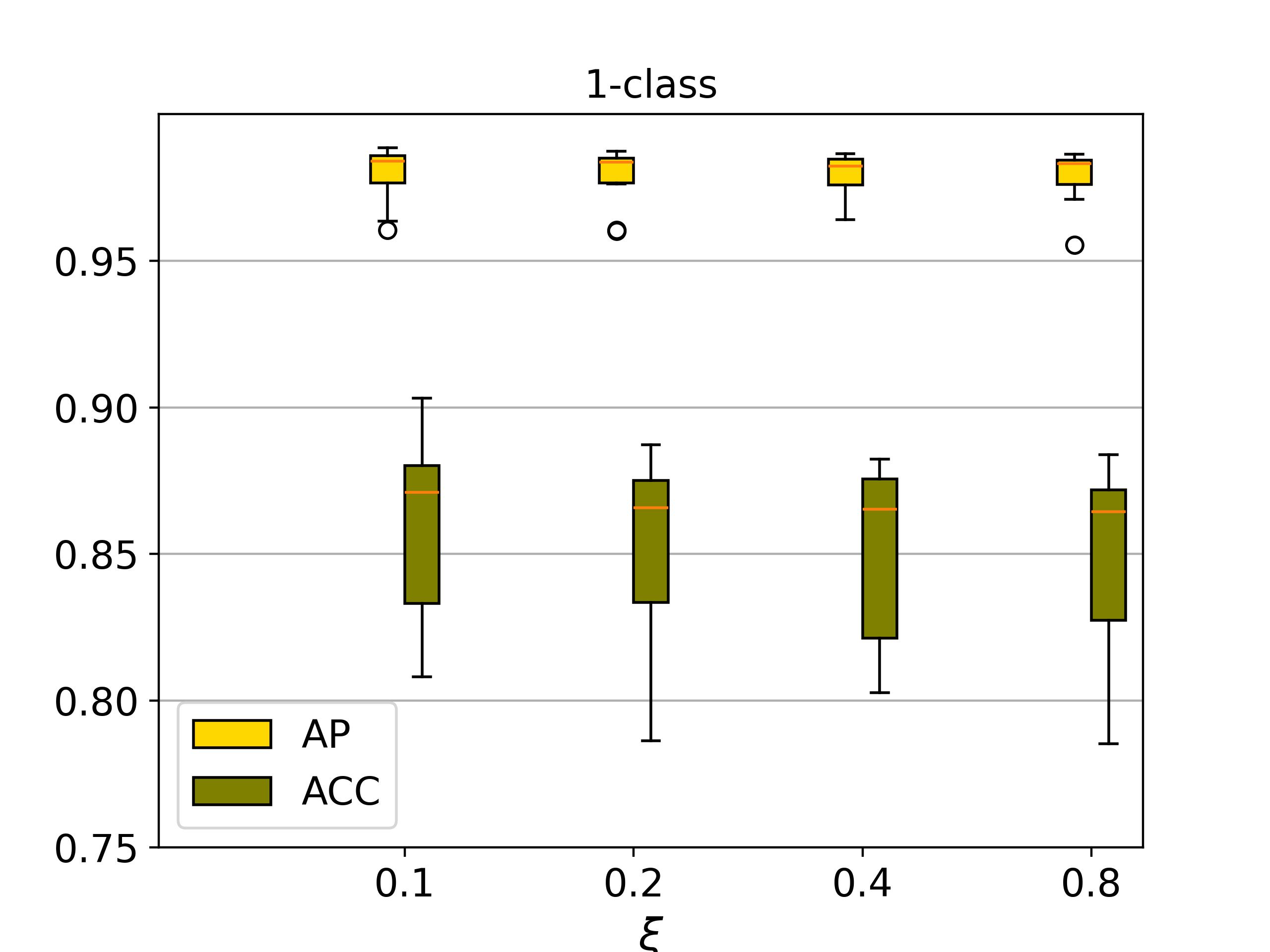}}
    \subfloat[]{\includegraphics[width=.33\linewidth]{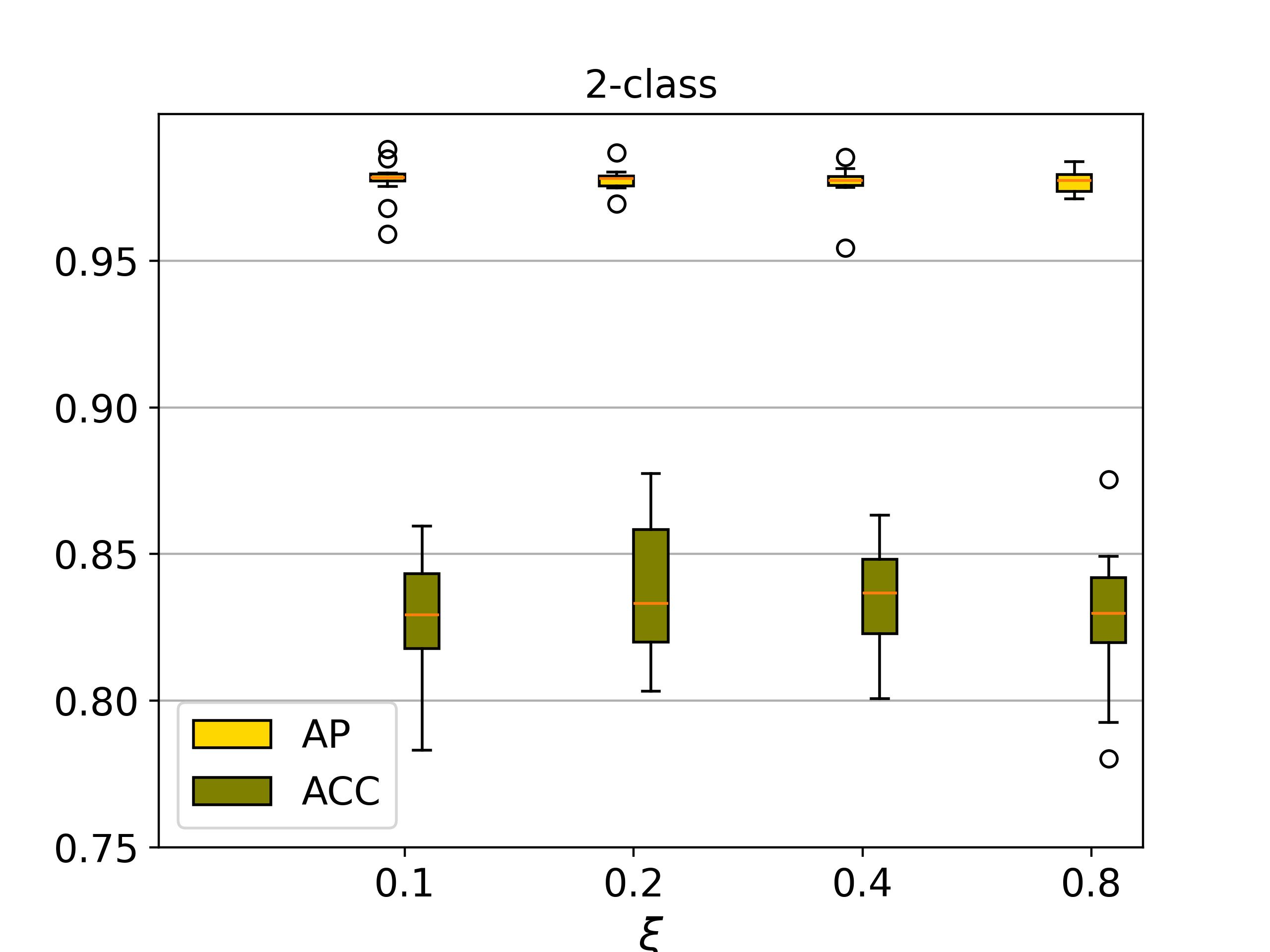}}
    \subfloat[]{\includegraphics[width=.33\linewidth]{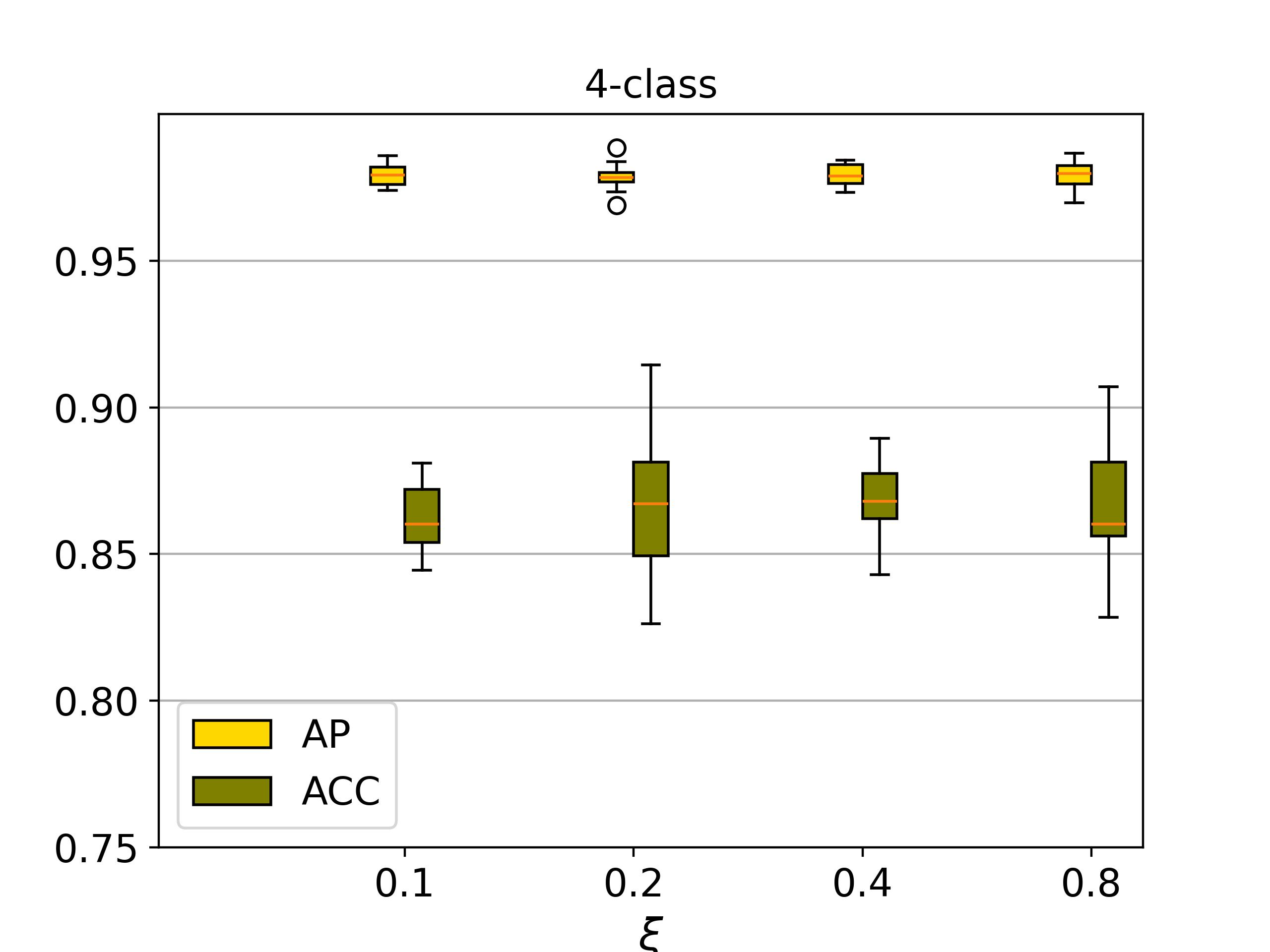}}
  
    \caption{Boxplots of ACC and AP expressing the impact of image encoder ViT-B/32 vs. ViT-L/14 for the (a) 1-class, (b) 2-class, and (c) 4-class settings, and impact of the contrastive loss factor $\xi$ for the (d) 1-class, (e) 2-class, and (f) 4-class settings.}
    \label{fig:hyperparams1}
\end{figure}

\Cref{fig:hyperparams1}(a)-(c) illustrates ACC \& AP boxplots for ViT-B/32 vs. ViT-L/14 in the 1-, 2-, and 4-class settings. Each boxplot is built from 48 scores (ACC/AP) obtained from the experiments of all combinations of $\xi$, $q$, and $d'$. It is clear that CLIP ViT-L/14 outperforms ViT-B/32, thus from now on we report results of ViT-L/14 only. \Cref{fig:hyperparams1}(d)-(f) illustrates the impact of the contrastive loss factor $\xi$, showing little variability across the different values, in all training settings.
\begin{figure}[t]
    \centering
    \includegraphics[width=0.8\linewidth]{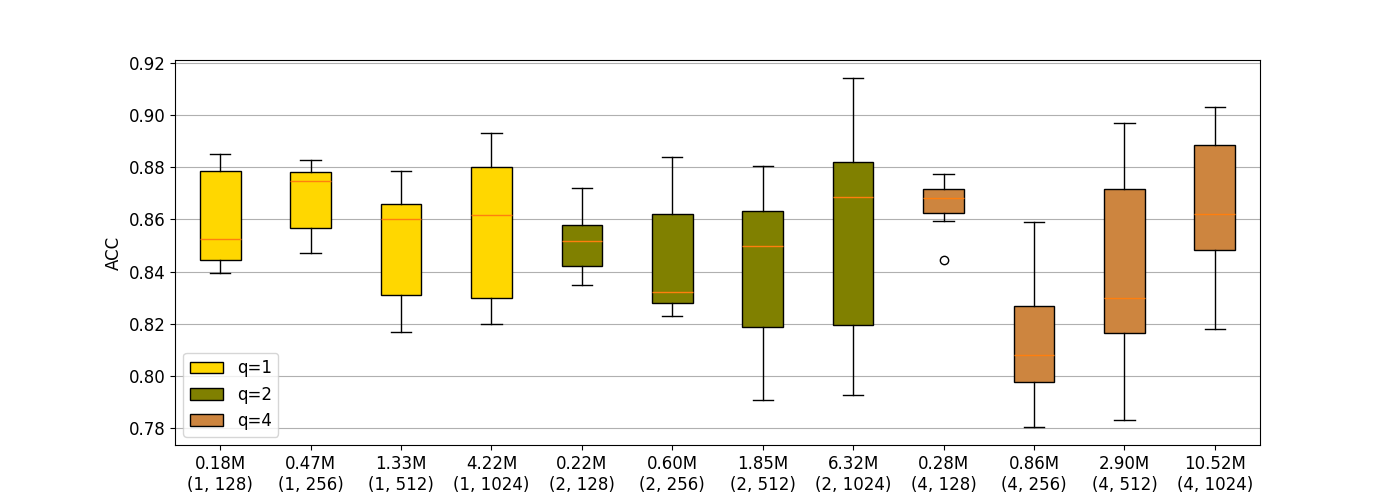}
    \caption{Boxplots of ACC for all different architectures of the trainable part of the model. X-axis shows the number of trainable parameters as well as the tuple ($q$, $d'$).}
    \label{fig:hyperparams2}
\end{figure}
\Cref{fig:hyperparams2} presents ACC boxplots for each ($q$, $d'$) combination of the trainable part of the model. For $q$=1, little variability is observed for different $d'$ choices, while for $q$=2 and $q$=4, there is a slight performance increase with $d'$ on average.  
The hyperparameter choices we have made for the proposed models are based on the maximum ACC \& AP and not their average or distribution.

\subsection{The Effect of Training Duration}
\label{subsec:training_duration}
\begin{figure}
    \centering
    
    \subfloat[]{\includegraphics[width=.33\linewidth]{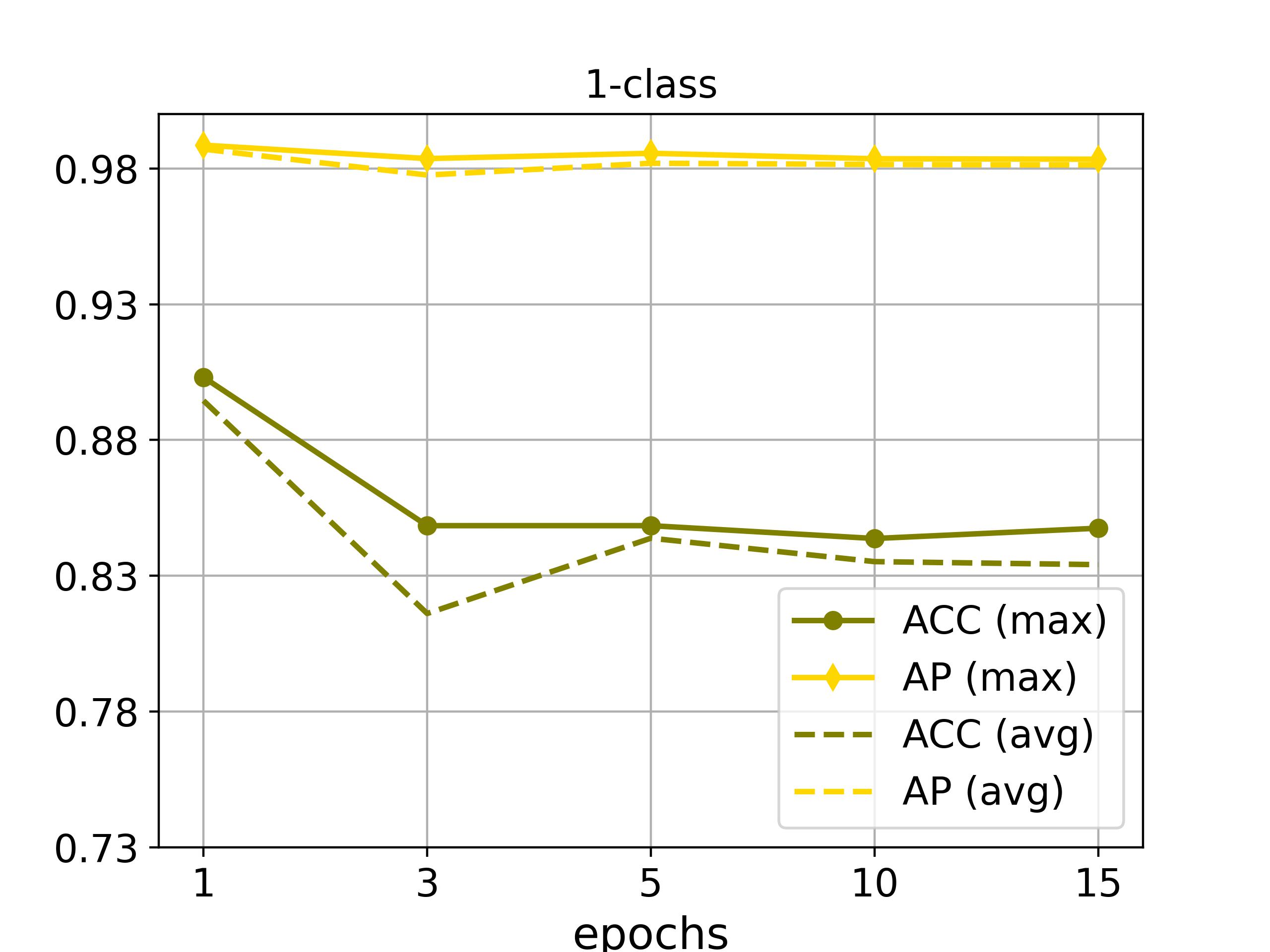}}
    \subfloat[]{\includegraphics[width=.33\linewidth]{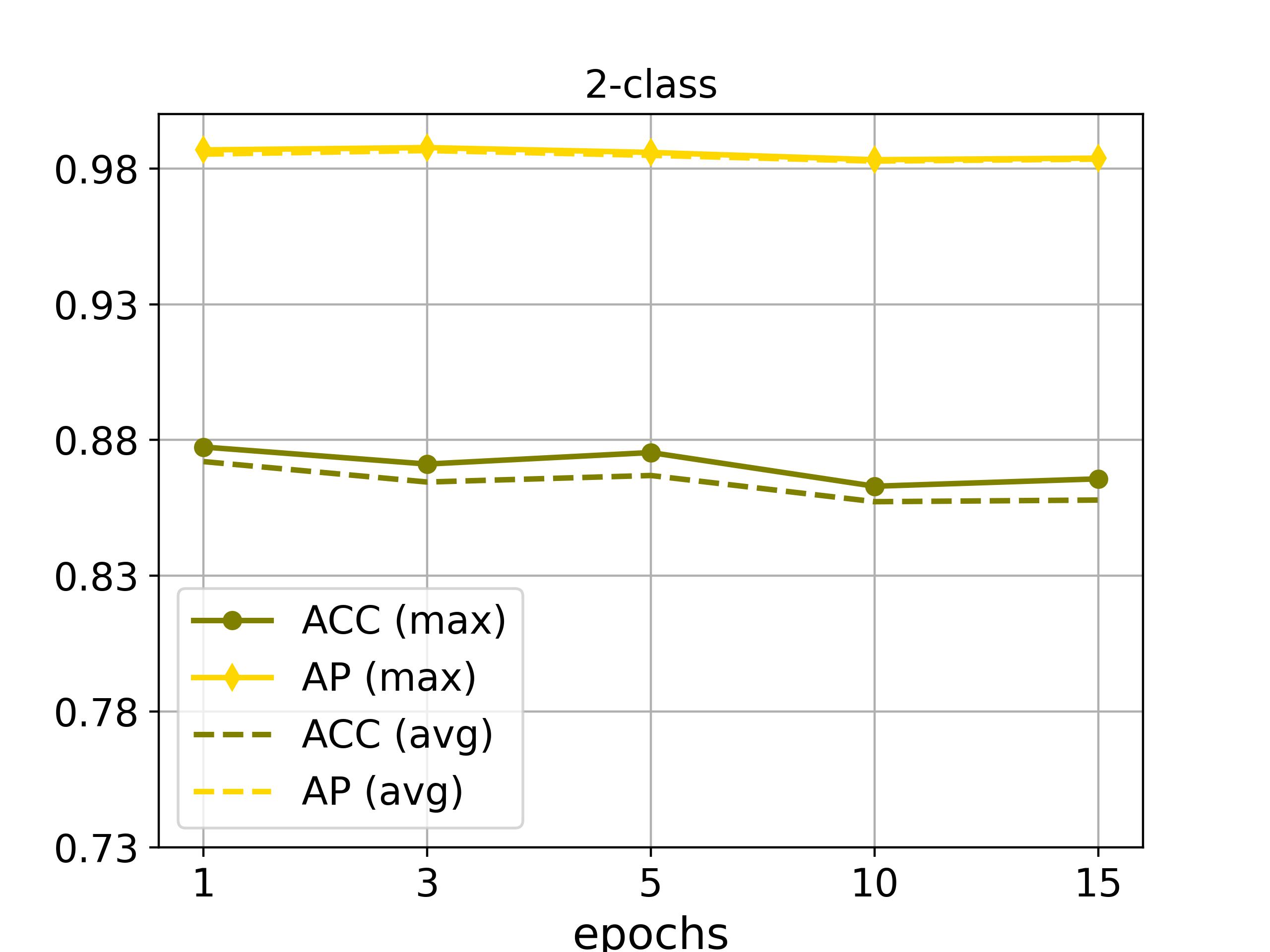}}
    \subfloat[]{\includegraphics[width=.33\linewidth]{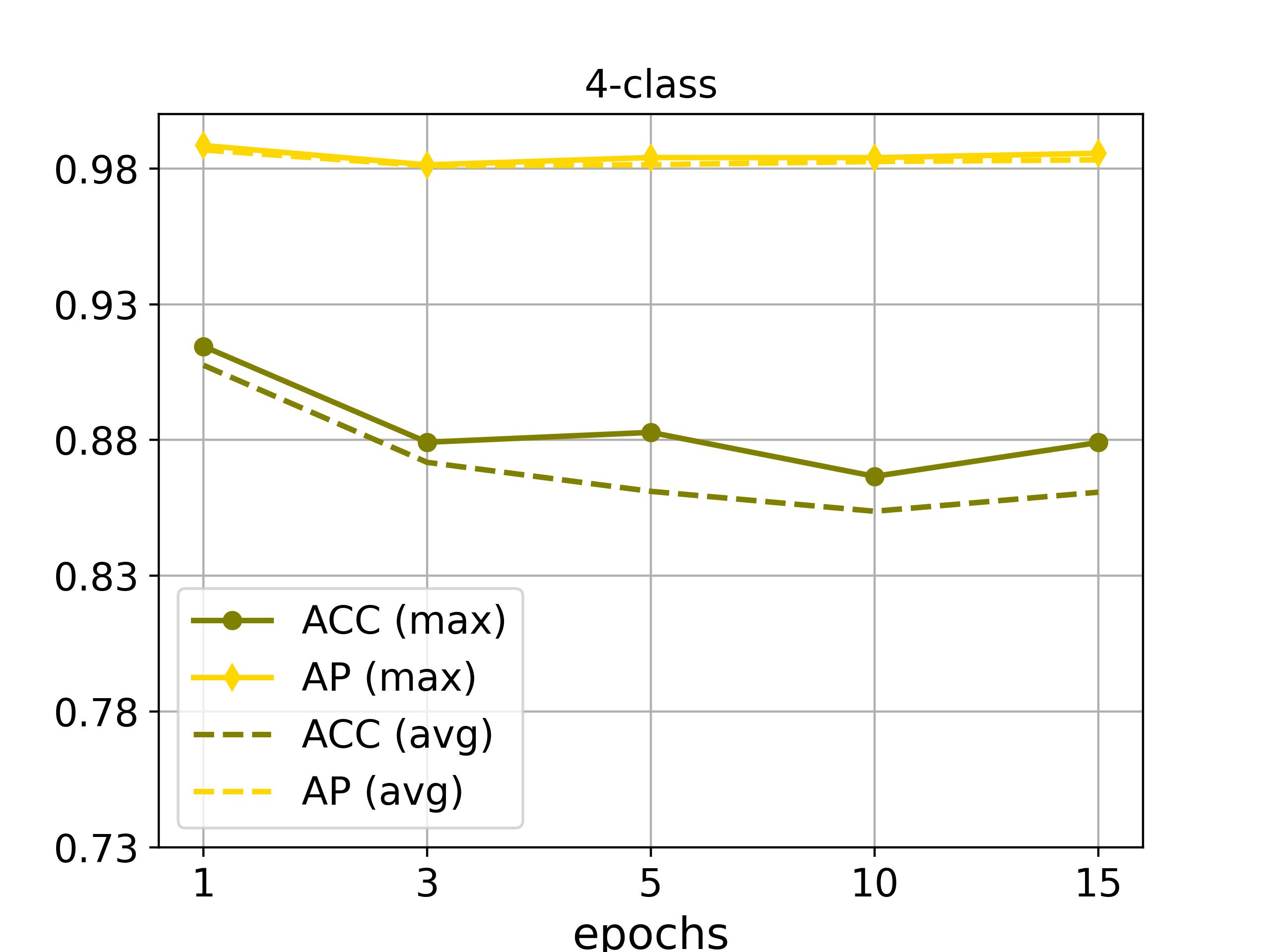}}
  
    \caption{Maximum and average performance (ACC \& AP) across the 20 evaluation datasets, of the 3 best-performing configurations at 3, 5, 10, and 15 epochs, for the (a) 1-class, (b) 2-class, and (c) 4-class training settings.}
    \label{fig:epochs}
\end{figure}

As stated in \cref{subsec:implementation}, the proposed models are trained for only one epoch. Here, we experimentally assess the potential benefit from further training. We identify the 3 best-performing configurations in the 1-epoch case for each of the 1-, 2-, and 4-class settings. Then, we train the corresponding models for 15 epochs and evaluate them in the 3rd, 5th, 10th, and 15th epoch (reducing learning rate by a factor of 10 at 6th and 11th epoch). \Cref{fig:epochs} presents their average and maximum performance, across the 20 test datasets, at different training stages. Neither the average nor the maximum performance improve if we continue training; instead, in the 1- and 4-class training settings performance drops.

\subsection{The Effect of Training Set Size}
\begin{figure}
    \centering
    
    \subfloat[]{\includegraphics[width=0.75\textwidth]{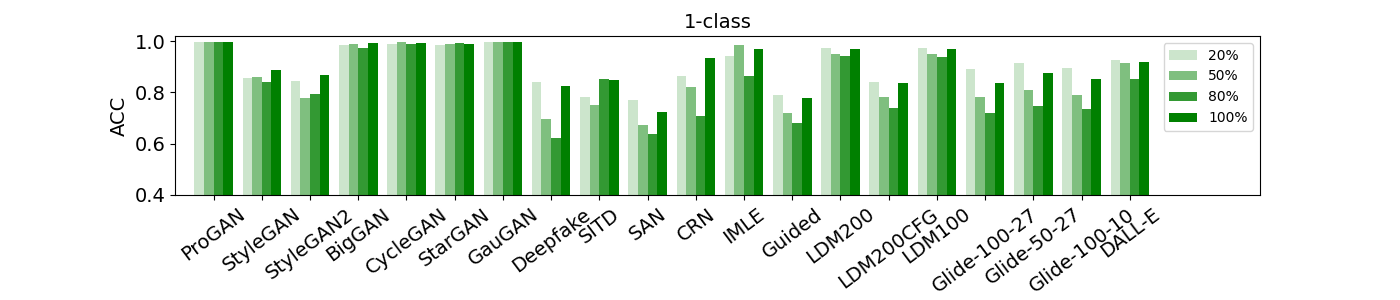}}
    \newline
    \subfloat[]{\includegraphics[width=0.75\textwidth]{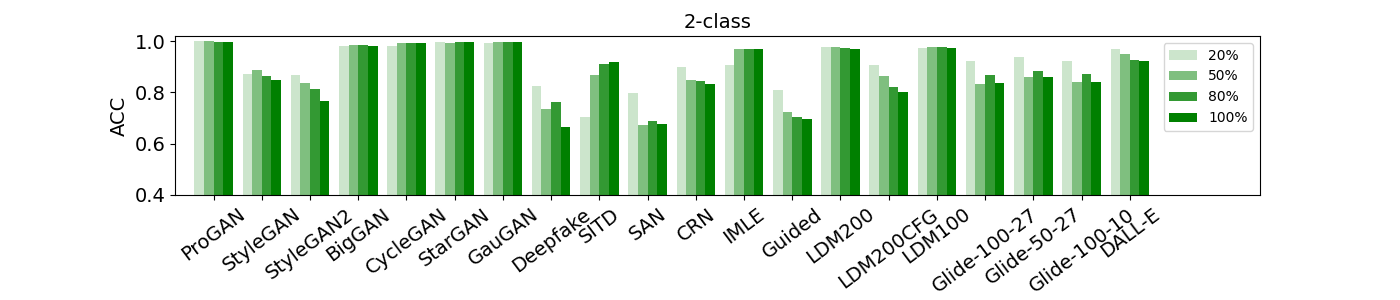}}
    \newline
    \subfloat[]{\includegraphics[width=0.75\textwidth]{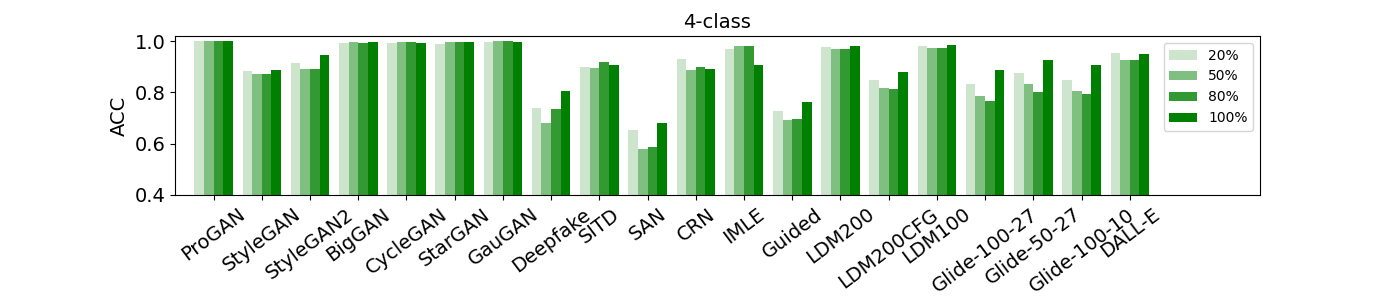}}
    \newline
    \subfloat[]{\includegraphics[width=0.75\textwidth]{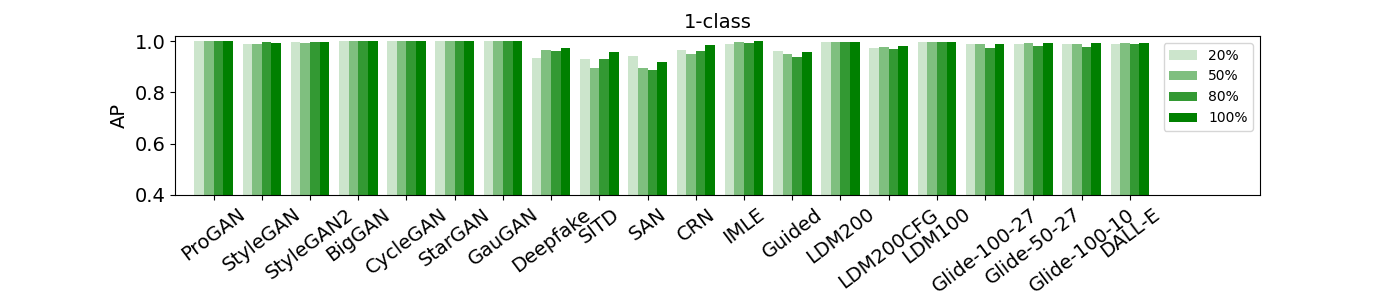}}
    \newline
    \subfloat[]{\includegraphics[width=0.75\textwidth]{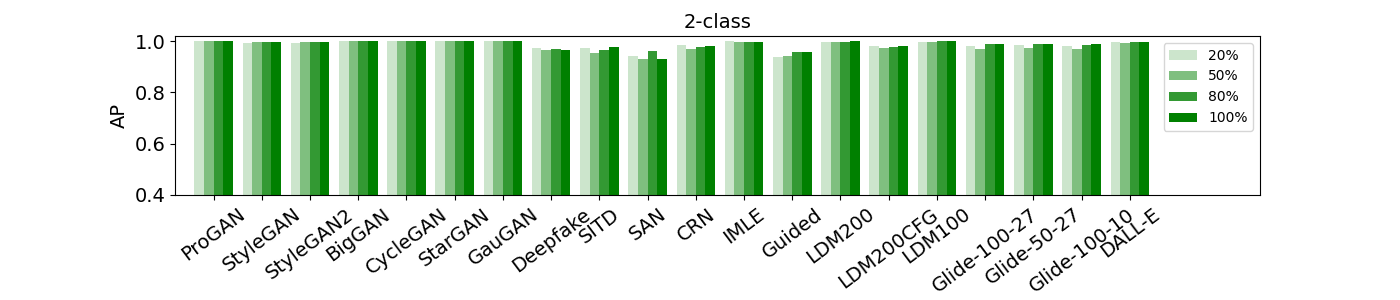}}
    \newline
    \subfloat[]{\includegraphics[width=0.75\textwidth]{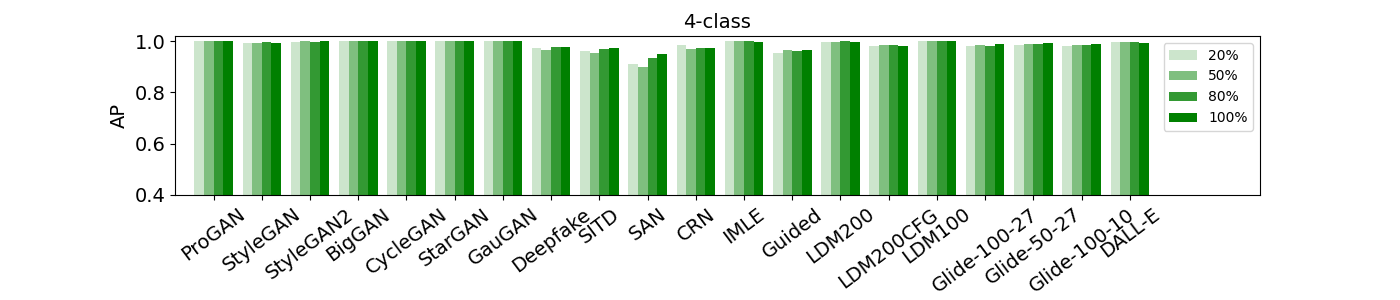}}
    \newline
  
    \caption{ACC on the 20 evaluation datasets when the proposed model is trained with 20\%, 50\%, 80\%, and 100\% of the training data, for the (a) 1-class, (b) 2-class, and (c) 4-class settings. In (d)-(f) the same for AP.}
    \label{fig:data_size}
\end{figure}

Here, we present a set of experiments in order to assess the effect of the training set size to our models' performance. Specifically, we re-train (again for only one epoch) the best models (1-class, 2-class, 4-class) using 20\%, 50\%, 80\%, and 100\% of the training data. For the 1-class, these percentages correspond to 7K, 18K, 28K, and 36K images, for the 2-class to 14K, 36K, 57K, and 72K images, and for the 4-class to 28K, 72K, 115K, and 144K images. In \cref{fig:data_size}, we present the obtained performance (ACC \& AP) on the 20 test datasets. We observe that the results are very close to each other, especially in terms of AP, showcasing the effectiveness of the proposed method even in limited training data settings.

\subsection{Robustness to Common Image Perturbations}\label{subsec:robustness}
Following common practice \cite{yu2019attributing,wang2020cnn,tan2023learning, ojha2023towards}, we assess our method's robustness to perturbations typically applied in social media by applying blurring, cropping, compression, noise addition, and their combination to each evaluation sample with 0.5 probability (experimental setting directly taken from \cite{frank2020leveraging}). The results in \cref{tab:perturbations} indicate high robustness against cropping and compression, with minimal performance losses, and a good level of robustness against blurring and noise addition. Their combination considerably reduces performance but at similar levels to that of current SotA without applying perturbations (cf. \cref{tab:comparative_accuracy}).
\begin{table}[]
\caption{Performance (ACC \& AP) after applying common image perturbations.}
    \label{tab:perturbations}
    \resizebox{\textwidth}{!}{\begin{tabular}{cccccccccccccccccccccccc}
    \toprule
        &\multicolumn{7}{c}{Generative Adversarial Networks}&&\multicolumn{2}{c}{Low level vision}&&\multicolumn{2}{c}{Perceptual loss}&&\multicolumn{3}{c}{Latent Diffusion}&&\multicolumn{3}{c}{Glide}\\
        \cmidrule{2-8}\cmidrule{10-11}\cmidrule{13-14}\cmidrule{16-18}\cmidrule{20-22}\\
        & Pro- & Style- & Style- & Big- & Cycle- & Star- & Gau- & Deep- &&&&&&&200&200&100&&100&50&100&&AVG\\
        perturb.&GAN&GAN&GAN2&GAN&GAN&GAN&GAN&fake&SITD&SAN&&CRN&IMLE&Guided&steps& CFG&steps&&27&27&10&DALL-E&\\
        \midrule\midrule
        \multicolumn{24}{c}{ACC}\\
        \midrule
        \midrule
        \multicolumn{24}{c}{1-class}\\
        \midrule
        blur&96.3&78.3&76.2&91.3&90.3&94.1&93.9&72.3&88.9&61.4&&83.8&86.5&77.1&93.3&78.5&93.2&&77.5&79.0&77.0&88.9&83.9\\
        crop&99.8&85.7&81.0&98.9&99.5&98.9&99.8&79.8&89.2&63.9&&91.8&94.1&74.9&96.5&81.3&96.5&&82.2&86.3&84.7&91.9&88.8\\
        compress&99.1&81.2&77.3&96.9&99.0&98.4&99.7&78.7&89.2&62.8&&88.6&96.1&72.1&94.4&76.3&94.2&&79.3&81.0&81.6&89.0&86.7\\
        noise&97.5&77.0&73.8&91.8&95.7&94.0&98.2&80.1&88.6&63.5&&80.1&89.8&78.1&93.7&75.4&94.2&&83.1&85.2&83.2&88.4&85.6\\
        combined&88.8&72.3&71.7&80.7&87.7&90.1&89.3&75.3&88.3&62.8&&76.3&80.0&80.2&87.4&72.2&88.4&&78.4&81.1&79.2&82.6&80.6\\
        
        \midrule
        no&99.8&88.7&86.9&99.1&99.4&98.8&99.7&82.7&84.7&72.4&&93.4&96.9&77.9&96.9&83.5&97.0&&83.8&87.4&85.4&91.9&90.3\\

        \midrule
        \multicolumn{24}{c}{2-class}\\
        \midrule
        blur&97.2&75.4&68.1&90.3&94.3&95.4&97.0&60.7&88.9&59.6&&80.9&91.3&72.0&92.7&74.0&91.8&&77.2&78.2&76.8&87.4&82.5\\
        crop&99.8&81.3&72.5&98.0&99.4&99.6&99.8&64.5&89.7&61.4&&85.3&96.9&67.2&97.0&79.2&97.2&&83.1&85.3&83.4&92.5&86.7\\
        compress&99.4&78.0&70.1&94.6&99.4&98.5&99.7&64.0&88.9&61.6&&82.2&96.7&66.2&94.5&74.4&94.0&&79.2&81.4&80.3&88.7&84.6\\
        noise&97.7&74.2&67.5&89.5&96.2&93.5&98.8&69.0&88.6&62.6&&73.3&88.4&71.7&92.8&71.0&91.6&&79.5&80.3&79.5&87.1&82.6\\
        combined&90.6&69.6&65.4&81.4&88.2&91.9&92.5&66.7&89.2&61.6&&75.5&88.3&74.7&87.6&68.2&86.3&&76.1&78.5&76.0&81.3&79.5\\
        \midrule
        no&99.8&84.9&76.7&98.3&99.4&99.6&99.9&66.7&91.9&67.8&&83.5&96.8&69.6&96.8&80.0&97.3&&83.6&86.0&84.1&92.3&87.7\\
        
        \midrule
        \multicolumn{24}{c}{4-class}\\
        \midrule
        blur&97.7&86.9&90.4&93.9&91.5&98.5&95.1&72.3&91.7&57.8&&81.0&81.6&77.6&92.8&81.2&92.3&&83.2&87.2&85.2&92.2&86.5\\
        crop&100.0&88.5&93.4&99.5&99.4&99.5&99.8&79.2&92.2&62.1&&87.7&88.3&74.9&98.0&88.3&98.4&&88.6&92.4&90.1&95.0&90.8\\
        compress&99.9&87.5&89.2&98.1&99.2&99.6&99.7&72.8&91.4&61.9&&88.0&93.0&70.6&95.2&79.8&95.9&&85.8&88.4&87.2&91.2&88.7\\
        noise&97.7&80.0&78.4&91.3&93.9&94.8&98.0&78.2&91.9&59.6&&71.9&79.8&74.5&92.5&74.8&92.7&&83.6&86.0&84.2&88.1&84.6\\
        combined&91.9&75.8&78.9&82.6&88.7&91.2&92.9&72.5&92.5&58.7&&77.5&85.0&79.1&89.1&73.4&89.2&&82.8&85.2&83.5&82.5&82.7\\
        \midrule
        no&100.0&88.9&94.5&99.6&99.3&99.5&99.8&80.6&90.6&68.3&&89.2&90.6&76.1&98.3&88.2&98.6&&88.9&92.6&90.7&95.0&91.5\\

        \midrule\midrule
        \multicolumn{24}{c}{AP}\\
        \midrule
        \midrule
        \multicolumn{24}{c}{1-class}\\
        \midrule
        blur&99.7&94.6&93.6&97.6&98.6&99.5&99.8&93.2&96.1&76.1&&94.1&98.9&91.0&98.6&91.2&98.6&&93.1&94.0&92.9&96.8&94.9\\
        crop&100.0&98.8&99.5&99.9&100.0&99.9&100.0&96.9&96.5&83.5&&98.2&99.9&95.3&99.8&98.0&99.8&&99.1&99.4&99.4&99.3&98.2\\
        compress&100.0&98.5&99.0&99.7&99.9&99.9&100.0&96.1&96.0&82.3&&95.9&99.6&94.0&99.7&96.2&99.7&&98.6&98.8&98.8&99.1&97.6\\
        noise&99.8&92.5&93.0&98.2&99.2&98.7&99.9&89.1&95.8&82.9&&90.6&96.9&95.0&99.0&92.6&99.1&&96.8&97.3&96.5&97.7&95.5\\
        combined&97.8&84.9&84.3&92.1&97.0&97.6&98.2&84.0&95.9&77.8&&87.2&94.6&92.2&95.7&84.0&96.0&&88.7&89.9&87.7&92.1&90.9\\
        
        \midrule
        no&100.0&99.1&99.7&99.9&100.0&100.0&100.0&97.4&95.8&91.9&&98.5&99.9&95.7&99.8&98.0&99.9&&98.9&99.3&99.1&99.3&98.6\\

        \midrule
        \multicolumn{24}{c}{2-class}\\
        \midrule
        blur&99.8&94.7&92.2&97.7&99.6&99.8&99.9&90.6&96.2&77.3&&90.7&98.2&91.2&98.8&91.3&98.5&&93.8&93.9&93.4&97.4&94.7\\
        crop&100.0&99.4&99.4&99.9&100.0&100.0&100.0&95.8&96.8&86.4&&97.4&99.7&94.7&99.9&98.1&99.9&&98.8&99.0&98.9&99.6&98.2\\
        compress&100.0&99.1&98.8&99.7&100.0&100.0&100.0&94.9&96.3&85.2&&96.5&99.7&94.1&99.7&96.0&99.7&&98.3&98.5&98.4&99.2&97.7\\
        noise&99.8&93.1&91.7&98.2&99.5&99.3&99.9&84.6&96.4&84.3&&91.7&98.0&93.7&99.2&91.7&99.2&&97.2&96.9&96.7&98.3&95.5\\
        combined&97.8&84.8&81.0&92.8&98.0&98.2&98.5&78.3&96.6&77.2&&87.3&96.2&90.4&96.5&84.1&95.8&&90.4&91.4&90.7&93.4&91.0\\
        \midrule
        no&100.0&99.5&99.6&99.9&100.0&100.0&100.0&96.4&97.5&93.1&&98.2&99.8&95.7&99.9&98.0&99.9&&98.9&99.0&98.8&99.6&98.7\\
        
        \midrule
        \multicolumn{24}{c}{4-class}\\
        \midrule
        blur&99.9&97.1&98.4&98.7&99.6&99.9&99.8&93.5&97.3&74.2&&95.1&99.5&90.5&98.4&93.1&98.1&&94.4&96.0&94.9&97.7&95.8\\
        crop&100.0&99.5&100.0&99.9&100.0&100.0&100.0&97.8&96.6&86.9&&97.5&99.7&95.6&99.8&98.3&99.9&&99.0&99.4&99.0&99.5&98.4\\
        compress&100.0&99.4&99.9&99.9&100.0&100.0&100.0&96.1&97.2&86.0&&95.4&99.5&94.3&99.6&96.0&99.6&&98.6&98.9&98.6&98.8&97.9\\
        noise&99.9&96.5&97.6&98.7&98.9&99.1&99.9&88.0&97.4&86.5&&82.7&91.5&94.5&99.0&93.0&98.8&&97.8&98.1&97.8&98.0&95.7\\
        combined&98.2&89.4&91.7&93.3&97.9&98.2&98.4&82.1&97.3&77.9&&88.1&94.6&93.3&96.6&87.6&96.3&&92.4&94.9&93.7&93.5&92.8\\
        \midrule
        no&100.0&99.4&100.0&99.9&100.0&100.0&100.0&97.9&97.2&94.9&&97.3&99.7&96.4&99.8&98.3&99.9&&98.8&99.3&98.9&99.3&98.8\\
        \bottomrule
    \end{tabular}}
    
\end{table}

\subsection{Importance of Intermediate Stages}
\begin{figure}[t]
    \centering
    \includegraphics[width=\linewidth]{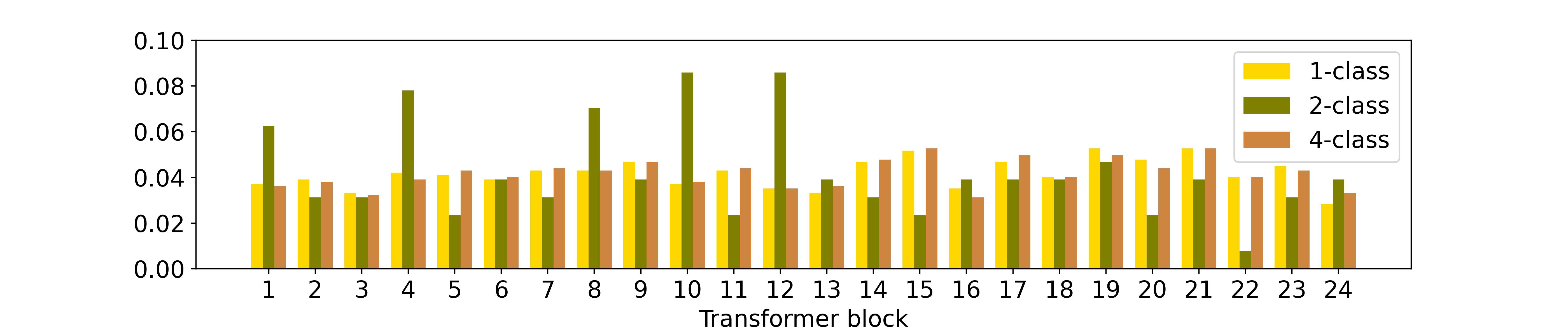}
    \caption{Frequency ($f_l$) of each Transformer block obtaining the maximum importance over feature elements.}
    \label{fig:importance}
\end{figure}

In \cref{subsec:ablations}, we show that the intermediate representations is the most important component of the proposed method. 
Here, we estimate the importance of each block by analysing the importances $\mathbf{A}=\{\alpha_{lk}\}$ estimated by TIE (cf. \cref{eq:tie}):
\[l^*_k=\underset{l}{\mathrm{argmax}} (\alpha_{lk})\text{, for $k$=1,\dots,$d'$}\]
\[f_l=\frac{\mid\{k\mid l^*_k=l\}\mid}{d'}\text{, for $l$=1,\dots,n}\]
\Cref{fig:importance} illustrates the frequency $f_l$ of Transformer block $l$ obtaining maximum importance, for $l$=1,\dots,$n$. For the 1-class and 4-class models $n$=24 and $d'$=1024, while for the 2-class model $n$=24 and $d'$=128. We observe that there are many Transformer blocks obtaining higher maximum importance frequency than the last one in all training settings, supporting the proposed method's motivation.

\subsection{The Effect of Training Data Origin on Detecting Diffusion Images}
We additionally assess our method's performance on Diffusion data produced by commercial tools, and more precisely on the Synthbuster dataset \cite{bammey2023synthbuster}. Due to the images' size we consider the ten-crop validation technique and average the predictions. \Cref{tab:synthbuster} illustrates RINE's performance on Synthbuster when trained on ProGAN data from \cite{wang2020cnn}, and on Latent Diffusion Model (LDM) data from \cite{corvi2023detection}. As expected training on Diffusion data provides big performance gains.

\begin{table}[h]
\caption{Performance (ACC/AP) of our method on Synthbuster.}
    \label{tab:synthbuster}
    \centering
    \resizebox{\textwidth}{!}{\begin{tabular}{c|ccccccccc}
    \toprule
                training set&DALL-E 2&DALL-E 3&SD1.3&SD1.4&SD2&SD XL&Glide&Firefly&Midjourney\\
                \midrule
        ProGAN (from \cite{wang2020cnn})&  64.6/86.7&21.1/30.7&66.3/91.3&65.8/91.4&49.0/66.8&53.3/73.4&42.6/55.7&70.8/99.3&34.2/39.5\\
        LDM (from \cite{corvi2023detection})& 89.8/96.2&47.2/32.5&96.4/100.0&96.4/100.0&93.5/98.5&96.3/99.8&90.0/96.3&85.3/93.4&92.4/97.4\\
        \bottomrule
    \end{tabular}}
    
\end{table}

\section{Conclusions}
\label{sec:conclusions}
In this work, inspired by the recent advances on the use of large-scale pre-trained visual features, we propose RINE that leverages representations from intermediate encoder-blocks of CLIP, as more effective to the SID task. Our comprehensive experimental study demonstrates the superiority of our approach compared to the state of the art. As an added benefit, the proposed approach requires only one epoch, translating to $\sim$8 minutes of training time, and limited training data to achieve maximum performance, while it is robust to common image transformations. 

\section*{Acknowledgments}
This work is partially funded by the Horizon Europe project vera.ai (GA No. 101070093) and H2020 project AI4Media (GA No. 951911).
\bibliographystyle{unsrt}  

\clearpage
\appendix
\section*{Appendix}

\begin{table}[h]
    \resizebox{\textwidth}{!}{
    \begin{tabular}{cccccccccccccccccccccccc}
    \toprule
        &\multicolumn{8}{c}{Generative Adversarial Networks}&\multicolumn{2}{c}{Low level vision}&&\multicolumn{2}{c}{Perceptual loss}&&\multicolumn{3}{c}{Latent Diffusion}&&\multicolumn{3}{c}{Glide}\\
        \cmidrule{2-8}\cmidrule{10-11}\cmidrule{13-14}\cmidrule{16-18}\cmidrule{20-22}\\
        & Pro- & Style- & Style- & Big- & Cycle- & Star- & Gau- & Deep- &&&&&&&200&200&100&&100&50&100&&AVG\\
        description&GAN&GAN&GAN2&GAN&GAN&GAN&GAN&fake&SITD&SAN&&CRN&IMLE&Guided&steps& CFG&steps&&27&27&10&DALL-E&\\
        \midrule
        \multicolumn{24}{c}{ACC}\\
        \midrule
        CLIP w/ FPN&99.1&81.8&75.0&90.0&86.1&68.4&99.3&55.1&72.8&50.7&&52.6&60.7&53.1&77.6&54.8&78.9&&58.2&58.4&57.6&61.9&69.6\\
        RN50 w/ FPN&89.7&69.3&63.7&58.4&79.9&62.3&77.1&52.2&70.3&60.3&&80.0&90.4&61.8&63.9&53.5&64.9&&64.5&65.0&64.2&52.8&67.2\\
        \midrule
        RINE 20-class&100.0&90.9&94.2&99.4&99.3&99.6&99.7&72.4&92.8&60.3&&92.7&96.9&73.8&98.5&94.7&98.7&&82.5&88.5&84.7&95.3&90.7\\
        UFD 4-class&99.6&82.0&72.7&94.8&99.1&96.4&99.4&71.4&63.6&56.2&&65.0&82.6&72.0&94.6&73.4&94.9&&75.5&75.9&74.8&87.4&81.6\\
        \midrule
        RINE w/ ViT&81.0&55.4&58.1&64.4&73.2&62.9&66.9&56.0&62.2&49.8&&66.5&67.5&54.9&56.0&47.3&56.3&&55.5&55.7&57.0&54.9&60.1\\
        RINE w/ Wang&85.6&63.7&60.0&58.8&77.5&62.5&73.5&52.3&76.1&56.2&&61.6&79.2&59.4&65.4&54.0&65.6&&63.6&63.2&62.8&52.8&64.7\\
        \midrule
         RINE 4-cl. (ours)&100.0&88.9&94.5&99.6&99.3&99.5&99.8&80.6&90.6&68.3&&89.2&90.6&76.1&98.3&88.2&98.6&&88.9&92.6&90.7&95.0&91.5\\
        \midrule
        \multicolumn{24}{c}{AP}\\
        \midrule
        CLIP w/ FPN&100.0&99.4&98.6&99.8&99.3&99.7&100.0&93.8&84.7&85.7&&99.1&99.9&90.2&99.8&97.1&99.9&&98.9&98.8&98.6&98.5&97.1\\
        RN50 w/ FPN&96.8&79.6&73.6&65.1&85.5&87.3&88.1&56.8&82.2&70.8&&95.2&98.1&69.6&73.1&55.8&74.1&&70.9&72.5&71.7&54.2&76.0\\
        \midrule
        RINE 20-class&100.0&99.8&100.0&100.0&100.0&100.0&100.0&97.5&97.8&84.2&&98.1&99.9&95.4&99.9&99.3&99.9&&96.7&98.1&96.9&99.4&98.1\\
        UFD 4-class&100.0&96.7&98.7&99.3&99.9&99.7&100.0&85.5&64.0&76.7&&94.6&99.0&88.0&99.4&92.3&99.3&&94.1&94.5&93.8&97.7&93.6\\
        \midrule
        RINE w/ ViT&91.0&57.6&64.0&71.4&83.3&71.8&76.2&59.4&77.1&52.2&&68.8&72.1&56.9&57.5&45.7&58.3&&55.1&55.7&56.4&54.5&64.3\\
        RINE w/ Wang&94.5&76.8&73.6&66.3&85.1&87.1&85.6&58.2&80.3&60.0&&87.2&96.3&65.9&77.9&56.7&77.4&&76.3&77.5&75.6&57.0&75.8\\
        \midrule
         RINE 4-cl. (ours)&100.0&99.4&100.0&99.9&100.0&100.0&100.0&97.9&97.2&94.9&&97.3&99.7&96.4&99.8&98.3&99.9&&98.8&99.3&98.9&99.3&98.8\\
    \bottomrule
    \end{tabular}
    }
    \caption{Further experimental results. Accuracy (ACC) and average precision (AP) are reported.}\label{tab:results}
\end{table}

\textbf{Feature fusion using Feature Pyramid Networks (FPN) \cite{lin2017feature}.} Although RINE's novelty lies in the use of \textit{intermediate} CLIP features, rather than the (intentionally simple) fusion mechanism, 
we perform two experiments to assess the effectiveness of more complex fusion mechanisms such as that of Feature Pyramid Networks. 
In the first experiment (``CLIP w/ FPN'' in Table \ref{tab:results}
) we replace $\mathfrak{Q}_1$, $\mathfrak{Q}_2$, and TIE modules by an FPN. 
In the second (``RN50 w/ FPN'' in Table \ref{tab:results}
) we train a ResNet50 (pre-trained on ImageNet) with FPN. 
Incorporating FPN increases GPU memory consumption from 7GB to 28GB, and training time from 8min to 36min (1 epoch). 
Results are a lot worse than RINE in terms of ACC (69.6 vs. 91.5) and a little worse in terms of AP (97.1 vs. 98.8). 
Training of ResNet50 with FPN converges after 10 epochs. 
It consumes 5GB of GPU memory during training and needs 3min/epoch (31min in total). However, it still results in worse performance than RINE. 

\textbf{Fair comparison with UFD \cite{ojha2023towards}.} 
One could argue that the performance gain of RINE compared to UFD may be questionable, since training with more classes of ProGAN-generated images may make the model overfit to GAN-generated images and hurt its generalization capabilities. Thus, we additionally trained RINE on 20 classes (kept the 4-class configuration; performed no tuning), and UFD on 4 classes. 
The results are provided in Table \ref{tab:results} at lines ``RINE 20-class'' and ``UFD 4-class''. 
RINE 20-class roughly preserves its 4-class instance performance (90.7 ACC, 98.1 AP) without any further tuning. 
No overfitting on GANs is observed. 
The frozen CLIP features are likely robust enough to prevent a lightweight MLP from overfitting. 
UFD maintains its performance in the 4-class setting as well, exhibiting no significant performance increase. 

\textbf{Considering simpler backbones.} In order to assess the importance of the backbone choice (CLIP in RINE's case) we perform two experiments using the ImageNet-pretrained ViT and Wang's detector as backbones, respectively.
Table~\ref{tab:results} presents the results at lines ``RINE w/ ViT'' and ``RINE w/ Wang''. The performance significantly decreases with these backbone choices.

\end{document}